%% file: main.tex
\documentclass{article} 
\usepackage{iclr2026_conference,times}

\input{math_commands.tex}

\usepackage[utf8]{inputenc} 
\usepackage[T1]{fontenc}    
\usepackage{hyperref}       
\usepackage{url}            
\usepackage{booktabs}       
\usepackage{amsfonts}       
\usepackage{nicefrac}       
\usepackage{microtype}      
\usepackage{xcolor}         
\usepackage{hyperref}
\usepackage{url}
\usepackage{amsmath}
\usepackage[pdftex]{graphicx}
\usepackage{multirow}
\usepackage{array}
\usepackage{colortbl}
\usepackage{subcaption}
\usepackage{natbib}        

\definecolor{citecolor}{HTML}{0071bc}
\definecolor{paleplum}{rgb}{0.8, 0.6, 0.8}
\hypersetup{
  colorlinks,
  citecolor=citecolor,
  linkcolor=red
}
    
\title{MoNE: Replacing Redundant Experts with Lightweight Novices for Structured Pruning of MoE}

\author{Geng~Zhang\thanks{
Equal contribution. $^\dagger$Corresponding author.
}$^{\,\,\,}$, 
Yuxuan~Han$^*$, Yuxuan~Lou, Yiqi~Zhang, Wangbo~Zhao$^\dagger$, Yang~You$^\dagger$ \\
National University of Singapore \\
\texttt{\{zhangg,youy\}@comp.nus.edu.sg} \quad \texttt{wangbo.zhao96@gmail.com} \\
\texttt{\{han\_yuxuan,yuxuanlou,yiqi.zhang\}@u.nus.edu}
}

\iclrfinalcopy 
\begin{document}

\maketitle

\begin{abstract}
Mixture-of-Experts (MoE) enables efficient scaling of large language models by activating only a subset of experts per input token.
However, deploying MoE-based models incurs significant memory overhead due to the need to retain all experts in memory. 
While structured pruning is promising to reduce memory costs, existing methods often show suboptimal performance and unstable degradation in three dimensions: model architectures, calibration data sources, and calibration sample sizes.
This paper proposes \textbf{M}ixture-\textbf{o}f-\textbf{N}ovices-and-\textbf{E}xperts (\textbf{MoNE}), a novel expert pruning method that replaces redundant experts with lightweight novices to achieve effective and robust model compression. 
MoNE evaluates expert redundancy based on two metrics: access frequency and output variance. 
Experts exhibiting low usage and stable outputs are pruned and replaced with lightweight novices—unbiased estimations of their original outputs—minimizing performance degradation. 
Extensive experiments demonstrate that MoNE consistently outperforms baseline methods with minimal accuracy degradation across the three dimensions, confirming its effectiveness and robustness. 
Notably, it outperforms baselines by up to 2.72 for the average zero shot accuracy across nine downstream tasks under 25\% pruning ratio, with only 0.14 performance drop for Qwen2-57B-A14B. 
The code is available at \url{https://github.com/zxgx/mode-pd}.
\end{abstract}

\input{sections/intro}
\input{sections/background}
\input{sections/preliminaries}
\input{sections/method}
\input{sections/evaluation}
\input{sections/conclusion}

\section*{Acknowledgment}
We would like to acknowledge that computational work involved in this research work is supported by NUS IT’s Research Computing group using grant numbers CFP02-CF-004.
Yang You's research group is being sponsored by NUS startup grant (Presidential Young Professorship), Singapore MOE Tier-1 grant, ByteDance grant, NUS ARTIC grant, Apple grant, Alibaba grant, Google Research and Google grant for TPU usage.

\bibliography{ref}
\bibliographystyle{iclr2026/iclr2026_conference}

\appendix
\input{sections/appendix}

\end{document}

%% file: math_commands.tex
\usepackage{amsmath,amsfonts,bm}

\def\eqref#1{equation~\ref{#1}}

\def\1{\bm{1}}

\DeclareMathAlphabet{\mathsfit}{\encodingdefault}{\sfdefault}{m}{sl}
\SetMathAlphabet{\mathsfit}{bold}{\encodingdefault}{\sfdefault}{bx}{n}



%% file: sections/intro.tex
\section{Introduction}
Mixture-of-Experts (MoE) has emerged as a powerful architecture for advancing the capabilities of large language models (LLMs) \citep{liu2024deepseek, liu2025muon, muennighoff2024olmoe}.
MoE-based LLMs achieve higher parameter efficiency than vanilla transformer-based LLMs by replacing the MLP module with a set of smaller MLP modules (experts) and sparsely activating partial experts for each input token \citep{lepikhin2021gshard}.
Despite its performance benefits, the deployment of MoE-based models often incur additional memory overhead to maintain the non-activated experts in memory, which is valuable but limited for existing accelerators such as GPU and TPU \citep{jouppi2023tpu}.

\begin{figure}[t]
    \centering
    \includegraphics[width=\linewidth]{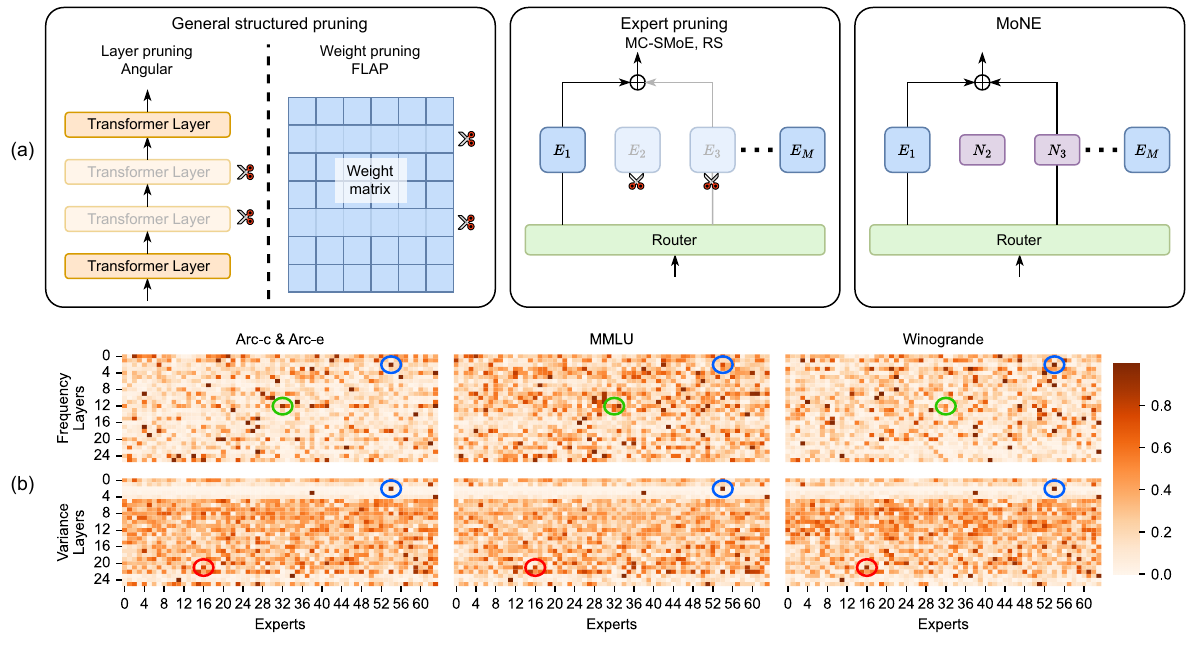}
    \caption{
        (a) Different structured pruning methods. 
        (b) Layer-wise normalized expert access frequency and output variance of Deepseek-V2-Lite for three downstream tasks. 
        Experts with high access frequency or output variances are the same across downstream tasks.
        Expert in blue circles \textcolor[HTML]{005FFF}{has both high frequency and variance}. Expert in red circles \textcolor[HTML]{FF0000}{only has high variance}. Expert in green circles \textcolor[HTML]{28C800}{only has high frequency}. 
        Similar observations on other models and tasks are in Appendix \ref{redundant experts}.
        }
    \label{intro}
\end{figure}

While diverse structured pruning methods have been proposed to reduce deployment memory costs by removing different model components while minimizing the performance degradation \citep{voita-etal-2019-analyzing, he2024demystifying, xia2024sheared, zhao2026dydit, dynamic, zhao2025rapid}, we observe that these approaches often exhibit \textit{suboptimal performance and unstable degradation} when applied to different MoE models.
Specifically, we identify three critical dimensions where existing methods fall short: model architectures, calibration data sources and calibration sample sizes, as shown by experiments in Section \ref{calib size}.
These limitations are evident across two main categories of structured pruning approaches for MoE models: general structured pruning and expert pruning as shown in Figure \ref{intro} (a).
First, general structured pruning methods that remove model layers (Angular \citep{gromov2025the}) or weight matrix channels (FLAP \citep{an2024fluctuation}) fail to account for the sparse computation scheme of MoE models when evaluating model component importance, resulting in inconsistent performance drop across the aforementioned three dimensions.
Second, existing expert pruning methods such as MC-SMoE \citep{li2024merge} and RS \citep{he2024demystifying} remove experts from MoE models primarily based on the expert access frequency.
However, as shown in Figure \ref{intro} (b), this feature alone fails to fully capture the expert redundancy. 
Besides, these methods lack mechanisms to recover the performance loss caused by pruning.

To improve the effectiveness and robustness of structured pruning for MoE models, this paper proposes a novel expert pruning method, \textbf{M}ixture-\textbf{o}f-\textbf{N}ovices-and-\textbf{E}xperts (\textbf{MoNE}) which replaces redundant experts with a lightweight structure, \textit{novice}. 
Specifically, to prune an MoE model with MoNE, it first evaluates the expert redundancy by the access frequency and the output variance for each expert on a calibration dataset.
Then, it identifies and prunes redundant experts that show low access frequency and stable output activations to reduce the memory overhead from redundant experts.
Finally, the unbiased estimation of the pruned expert output is employed as the lightweight novice to reclaim the performance loss caused by the pruned expert.
The intuition behind MoNE is that experts with low access frequency contribute less to the final outputs and experts whose outputs have low variance can be replaced with a constant but introduce less discrepancy.
Moreover, Figure \ref{intro} (b) reveals that experts with less redundancy identified by MoNE exhibit strong consistency across various downstream tasks.

The contribution of this paper is summarized as follows:
\begin{itemize}
    \item We propose a novel expert pruning method named MoNE which replaces redundant experts with lightweight novices to compress MoE models with minimal performance loss. 
    \item We exploit the expert access frequency and output variance to measure the expert redundancy and employ the unbiased estimation of the expert output to minimize the output discrepancy after pruning, thus achieving effective and robust pruning results.
    \item Extensive experiment results demonstrate that MoNE consistently outperforms baseline methods under varying MoE architectures, calibration data sources and calibration sample sizes. Notably, it outperforms baselines by up to 2.72 for the average zero shot accuracy across nine downstream tasks under 25\% pruning ratio, with only 0.14 performance drop for Qwen2-57B-A14B. 
\end{itemize}

%% file: sections/background.tex
\section{Related Work}
Model pruning compresses a model by removing certain redundant model parameters while preserving accuracy.
Existing pruning methods generally fall into two categories: \textit{unstructured pruning} and \textit{structured pruning}.
Unstructured pruning eliminates any model parameter that has minimal impact on model performance.
Methods such as SparseGPT \citep{frantar2023sparsegpt}, Wanda \citep{sun2024a}, SparseLLM \citep{bai2024sparsellm} excel in maintaining accuracy while achieving high compression ratios.
However, the resulting irregular sparsity patterns hinder efficient representation and execution on hardware accelerators.

In contrast, structured pruning removes certain modules of a model, preserving hardware-friendly structures.
Early researches prune redundant transformer layers of a LLM \citep{Fan2020Reducing, ling2024slimgpt, gromov2025the}.
LLM-Pruner \citep{ma2023llm}, FLAP \citep{an2024fluctuation}, MoE-Pruner \citep{xie2024moe} and SlimMoE \citep{li2025slimmoe} remove rows or columns of individual weight matrices.
Recent work also proposes to delete components such as attention, MLP or MoE modules within each transformer layer \citep{voita-etal-2019-analyzing, he2024demystifying}.
Minitron \citep{muralidharan2024compact} and Sheared LLaMA \citep{xia2024sheared} combine different granularity and automatically search for the optimal structures to prune. 
Despite their versatility, existing structured pruning methods often exhibit inconsistent performance across MoE architectures.

This work focuses on expert pruning, a unique direction of structured pruning for MoE models \citep{lepikhin2021gshard, liu2024deepseek}.
Expert pruning targets on deleting individual experts for each layer to compress an MoE model.
Previous expert pruning methods either require exhaustive search to identify redundant experts \citep{lu2024not}, or heavily rely on retraining to recover accuracy due to the suboptimal pruning performance \citep{li2024merge, he2024demystifying}.
However, the exhaustive search is not applicable to modern MoE model architectures such as Deepseek \citep{deepseekv2, liu2025muon}, OLMoE \citep{muennighoff2024olmoe} or Qwen \citep{team2024qwen2, yang2025qwen3}, as their MoE layer contains 64 experts or even more, yielding a tremendous search space that is intractable.
Retraining obscures the advantages of expert pruning over other structured pruning methods.

%% file: sections/preliminaries.tex
\section{Preliminaries}
\subsection{Mixture-of-Experts (MoE)}

MoE-based LLMs replace the traditional MLP module in the transformer layer with MoE module.
Each MoE module consists of a router network $G$ and a set of experts $\mathcal{E} = \{E_1, E_2, \dots, E_M\}$, where $M$ is the number of experts and each expert is a smaller MLP.
Let $\textbf{x} \in \mathbb{R}^d$ be the hidden state of an input token, where $d$ is the hidden size of the model, the output of an MoE module is computed as:
\begin{align} \label{moe definition}
\mathrm{MoE}(\textbf{x}, G, \mathcal{E}) = \sum_{E_i \in \mathcal{S}_{k,\textbf{x}}} G_i(\textbf{x}) \cdot E_i(\textbf{x})
\end{align}

The output of the router network $G(\textbf{x}) \in \mathbb{R}^M$ represents the routing scores for all experts, and $\mathcal{S}_{k,\textbf{x}} \subseteq \mathcal{E}$ denotes the top $k$ experts with the highest routing scores for input $\textbf{x}$.
The final output of the MoE module is the weighted sum of outputs from the top $k$ experts.
While Equation \ref{moe definition} captures the general MoE computation, implementations for $G$ and $E_i$ may vary across model architectures \citep{deepseekv2, liu2024deepseek, muennighoff2024olmoe, liu2025muon}.

\subsection{Expert pruning formulation}
Previous studies have revealed that not all experts contribute equally, and pruning less important ones can reduce memory overhead with marginal performance degradation \citep{lu2024not, li2024merge, huang2025mixture}.
However, searching for the target experts to prune at the global model perspective falls into a tremendous search space, as the number of experts per transformer layer increases with the evolving of the MoE model architectures \citep{lepikhin2021gshard, jiang2024mixtral, deepseekv2, liu2024deepseek}.
Following the layer-wise pruning scheme \citep{frantar2023sparsegpt, an2024fluctuation, lu2024not, ling2024slimgpt}, our goal of expert pruning is to identify a subset of redundant experts $\mathcal{P} \subseteq \mathcal{E}$ such that we can minimize the output difference after compressing their parameters:
\begin{align} \label{objective}
\underset{\mathcal{P} \subseteq \mathcal{E}}{\min} \Vert \mathrm{MoE}(\textbf{x}, G, \mathcal{E} \setminus \mathcal{P}) -  \mathrm{MoE}(\textbf{x}, G, \mathcal{E}) \Vert_2
\end{align}
To achieve this goal, the core problem is twofold: (1) find \textit{a metric to evaluate the importance of the experts in each layer}, so that we can identify the expert subset $\mathcal{P}$, and (2) find \textit{an pruning method to compress the parameters of $\mathcal{P}$}, so that we can reduce the model size.
While $\mathcal{E} \setminus \mathcal{P}$ implies directly removing redundant experts \citep{he2024demystifying, lu2024not}, existing methods have also explored expert merging to mitigate the expert redundancy \citep{li2024merge}.

%% file: sections/method.tex
\begin{figure}[t]
    \centering
    \includegraphics[width=\textwidth]{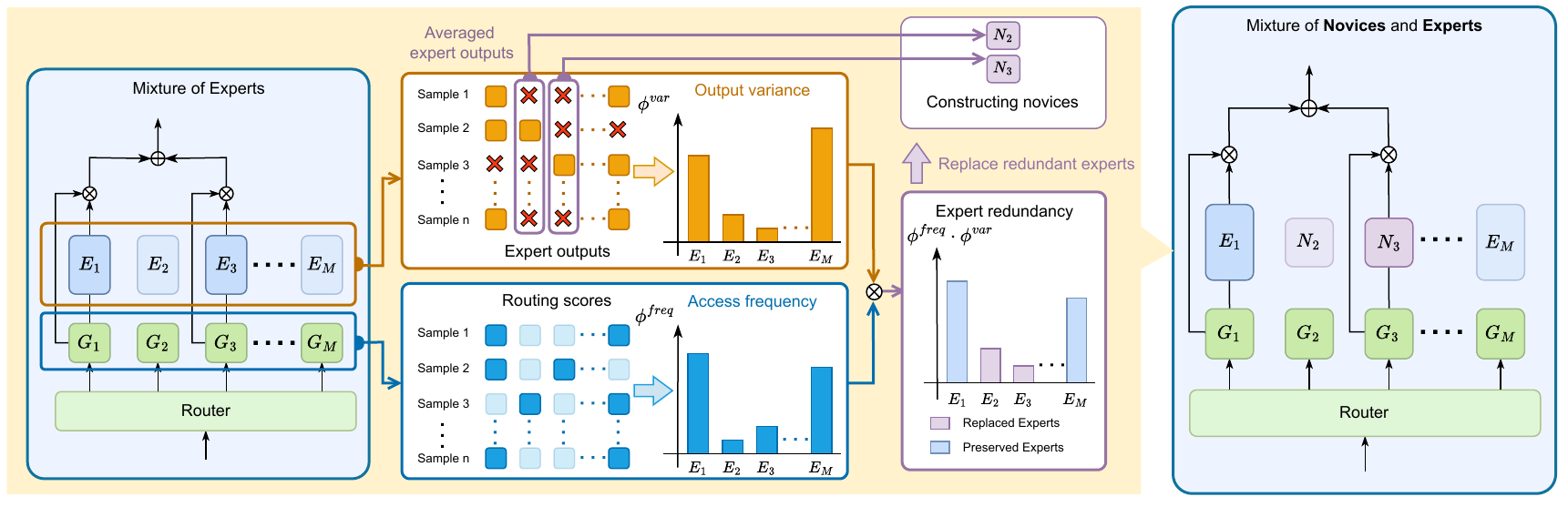}
    \caption{
        \textbf{The overview of MoNE.}
        Given an MoE model, it first exploits a calibration dataset to evaluate the expert access frequency and output variance.
        Then, the two metrics are fused to get the expert redundancy.
        Finally, the novices are derived from the averaged outputs for redundant experts.
    }
    \label{overview}
\end{figure}

\section{Mixture of Novices and Experts}
This section introduces \textbf{Mixture-of-Novices-and-Experts (MoNE)}, a novel expert pruning method designed to achieve effective and robust compression for MoE models while minimizing performance degradation.
Section \ref{framework} presents the computational framework of MoNE.
Section \ref{redundancy metric} defines the metric to evaluate the redundancy of experts.
Section \ref{prune process} explains the pruning process that compresses the redundant experts to lightweight novices.
The overview of MoNE is depicted in Figure \ref{overview}.

\subsection{MoNE framework} \label{framework}
MoE models are often trained with auxiliary losses to ensure load balance among experts in each layer, enabling each expert to learn certain aspect of knowledge \citep{lepikhin2021gshard, deepseekv2, muennighoff2024olmoe}.
However, most existing expert pruning methods directly remove experts \citep{he2024demystifying, li2024merge}, often leading to inconsistent performance drops across different model architectures or calibration data.
MoNE addresses this issue by introducing lightweight structures called \textbf{novices} to replace the pruned experts.
A novice is designed to capture the essential knowledge previously held by the removed experts. 
In contrast to simply removing redundant experts, MoNE compensates for knowledge loss by leveraging novices, thereby preserving the overall performance of the model while maintaining compression efficiency.
Specifically, the output of the MoNE is computed as:

\begin{align} \label{compute MoNE}
\mathrm{MoNE}(\textbf{x}) = \left(\sum_{E_i \in \mathcal{S}_{k,\textbf{x}} \setminus \mathcal{P}}G_i(\textbf{x})\cdot E_i(\textbf{x}) \right)+\left(\sum_{E_i \in \mathcal{S}_{k,\textbf{x}} \cap \mathcal{P}}G_i(\textbf{x})\cdot N_i \right)
\end{align}

where $\mathcal{S}_{k,\textbf{x}} \setminus \mathcal{P}$ and $\mathcal{S}_{k,\textbf{x}} \cap \mathcal{P}$ denote the preserved and pruned experts among the top $k$ activated experts respectively.
$N_i \in \mathbb{R}^d$ is the novice $i$ that retains the essential knowledge of the pruned expert.
Notably, $N_i$ is a compressed vector that does not involve any computation with the input token $\textbf{x}$. 
As a result, the computation and memory overhead is nearly identical to directly removing experts.
Furthermore, replacing experts with novices introduces adaptive computation overhead for different tokens, leading to fewer activated parameters for tokens routed to novices.
Nevertheless, empirical results in Section \ref{effect exp} demonstrate that MoE models pruned by MoNE maintain more zero shot performance on downstream tasks compared to existing expert pruning methods that only remove experts but keep the same activated parameters.

\subsection{Expert redundancy evaluation} \label{redundancy metric}
To identify the expert subset $\mathcal{P}$, we introduce an \textit{expert redundancy score} $\phi$ to assess the redundancy of experts.
To ensure the pruned experts contribute minimally to the model’s overall performance, the expert redundancy score $\phi$ takes two aspects into consideration: the variance in an expert’s output across a calibration dataset $\mathcal{C}$, and the frequency of an expert selected by the router network $G$.

\paragraph{Variance-based redundancy}
As the novices are constant vectors to ensure reduced computation and memory overhead, the outputs of the pruned experts are expected to have low variance across a calibration dataset $\mathcal{C}$.
In other words, experts with high output variance should be retained to contribute more discriminative information during inference, whereas experts with low output variance could be compressed into a more efficient representation, i.e., a novice.
The second row of Figure \ref{intro} (b) visualizes this motivation.
Expert outputs exhibit diverse variances, but we can find experts in blue and red circles that \textit{maintain high variances across different downstream tasks}.
Therefore, we introduce a variance-based redundancy $\phi^{var}_i$ to measure the output variance for expert $E_i$.
Concretely, $\phi^{var}_i$ is the L2 norm of the unbiased estimation for the output variance:
\begin{align}
\phi_i^{var} &= \left \| \sqrt{\frac{
    \sum_{\textbf{x} \in \mathcal{C}} (E_i(\textbf{x})-\overline{E_i})^2 \cdot \mathbb{I}(E_i \in \mathcal{S}_{k,\textbf{x}})
}{
    \sum_{\textbf{x} \in \mathcal{C}} \mathbb{I}(E_i \in \mathcal{S}_{k,\textbf{x}}) - 1
}}\right\|_2
\\
\overline{E_i} &= \frac{
    \sum_{\textbf{x} \in \mathcal{C}} E_i(\textbf{x}) \cdot \mathbb{I}(E_i \in \mathcal{S}_{k,\textbf{x}})
}{
    \sum_{\textbf{x} \in \mathcal{C}} \mathbb{I}(E_i \in \mathcal{S}_{k,\textbf{x}})
} \label{novice compute}
\end{align}
where $\mathbb{I}(E_i \in \mathcal{S}_{k,\textbf{x}})$ is the indicator function to show whether $E_i$ is among top $k$ experts for the input token $\textbf{x}$ of the calibration dataset $\mathcal{C}$.

\paragraph{Frequency-based redundancy}
The routing scores and access frequencies of the router network $G$ serve as strong indicators of the overall redundancy of an expert \citep{he2024demystifying, li2024merge}.
Intuitively, experts which are rarely selected or consistently assigned lower routing scores are likely to have a minimal impact on the model’s output. 
As shown in Figure \ref{intro} (b), we can identify typical experts in blue and green circles that show consistent high frequency over the three downstream tasks. Notably, the expert in green circles only has high frequency.
Therefore, \textit{the frequency and variance information can complement the discrepancy ignored by each other.}
Based on this observation, we define the frequency-based redundancy $\phi_i^{freq}$ of the expert $E_i$ as the average routing score across a calibration dataset $\mathcal{C}$ of which $E_i$ is among the top $k$ selected experts.
Formally, the frequency-based redundancy $\phi_i^{freq}$ is defined as:
\begin{align}
\phi_i^{freq} = \frac{\sum_{\textbf{x} \in \mathcal{C}} G_i(\textbf{x})\cdot \mathbb{I}(E_i \in \mathcal{S}_{k,\textbf{x}})}{\sum_{\textbf{x} \in \mathcal{C}}\mathbb{I}(E_i \in \mathcal{S}_{k,\textbf{x}})}
\end{align}

Finally, the two redundancy metrics are fused to obtain the expert redundancy score $\phi$:
\begin{align}
    \phi = \phi^{var} \cdot \phi^{freq}
\end{align}
A lower expert redundancy score $\phi_i$ indicates higher redundancy for expert $E_i$, making it a suitable candidate for pruning and replacement with a novice $N_i$.

\subsection{Expert replacement with novice} \label{prune process}
After identifying the pruned expert subset $\mathcal{P}$, we need to construct lightweight novices to replace them.
According to Equation \ref{objective}, the general objective for expert pruning is to minimize the discrepancy introduced by the removed expert outputs.
Since the output after applying MoNE is formulated as Equation \ref{compute MoNE}, the concrete objective for MoNE can be translated to:
\begin{align} \label{MoNE objective}
\underset{E_i \in \mathcal{P}}{\min} \sum_{\textbf{x} \in \mathcal{C}} (\Vert E_i(\textbf{x})-N_i \Vert_2)
\end{align}
Because $N_i$ is a constant vector, the optimal novice vector $N_i$ that best approximates the output of a pruned expert $E_i$ can be obtained in a closed form, i.e., $\overline{E_i}$ in Equation \ref{novice compute}.

To sum up, MoNE uses the unbiased estimations of mean expert outputs to replace experts that have the minimum output variance. 
As a result, MoNE achieve the goal that effectively and robustly compresses the MoE experts while minimizing performance degradation.

%% file: sections/evaluation.tex
\section{Evaluation}

\subsection{Experiment setup} \label{exp setup}
\paragraph{Base MoE models}
To validate the effectiveness and robustness of MoNE, we conducted structured pruning on five open source MoE models with diverse architectures and model scales: \textbf{OLMoE} \citep{muennighoff2024olmoe}, \textbf{Moonlight} \citep{liu2025muon}, \textbf{DeepSeek-V2-Lite} \citep{deepseekv2}, \textbf{Qwen2-57B-A14B} \citep{team2024qwen2} and \textbf{Qwen3-30B-A3B} \citep{yang2025qwen3}.
OLMoE has 7B parameters with 1B activated parameters per token.
Both Moonlight and Deepseek-V2-Lite have 16B parameters with 3B activated pamaters per token.
OLMoE and Moonlight represent SOTA MoE models at their respective scales.
To demonstrate scalability to larger architectures, we also consider the Qwen series: Qwen3-30B-A3B with 30B parameters and 3B activated per token, and Qwen2-57B-A14B with 57B parameters and 14B activated per token.
We chose the base version of the five models for experiments.

\paragraph{Baseline methods}
We selected structured pruning methods for different structures as baseline.
Notably, unless explicitly stated, we did not apply any weight update to compare the effect of pruning methods.
Specifically, for general structured pruning methods, we used \textbf{Angular} for layer pruning \citep{gromov2025the}, which evaluates the layer importance by the angular distance between the input activations for different layers, and we used \textbf{FLAP} for weight pruning \citep{an2024fluctuation}, which evaluates the channel importance by the fluctuation of the input activations and compensates the performance loss with the averaged output activations.
For expert pruning methods, we adopted the expert merging method in \textbf{MC-SMoE} \citep{li2024merge} for one of the expert pruning baselines.
Another expert pruning baseline is \textbf{RS} \citep{he2024demystifying}, which uses routing scores to evaluate the expert importance and discards less accessed ones.

\paragraph{Implementation details}
We tested two pruning ratios: 25\% and 50\%.
To demonstrate the robustness of MoNE to calibration data, we conducted experiments on two calibration data sources: Zyda2 \citep{zyphra_nvidia_2024} and C4 \citep{raffel2020exploring}.
Both datasets are constructed for LLM pretraining and C4 is commonly used as the calibration dataset for model compression \citep{ling2024slimgpt, frantar2023sparsegpt, gromov2025the, xia2024sheared}.
Besides, we also investigated the performance under three calibration sample sizes: 100, 500 and 1000 in Section \ref{calib size}.

\paragraph{Evaluation protocol}
Following previous researches \citep{ma2023llm, bai2024sparsellm, ling2024slimgpt, xia2024sheared, an2024fluctuation}, we adopted lm-evaluation-harness\footnote{\url{https://github.com/EleutherAI/lm-evaluation-harness}} \citep{eval-harness} to measure the zero shot accuracy and average results on nine downstream tasks: Arc-c and Arc-e \citep{clark2018think}, BoolQ \citep{clark2019boolq}, COPA \citep{roemmele2011choice}, MMLU \citep{hendrycks2021measuring}, OBQA \citep{mihaylov2018can}, PIQA \citep{bisk2020piqa}, RTE \citep{wang2019superglue} and Winogrande \citep{sakaguchi2021winogrande}.

Though more complex downstream tasks such as coding \citep{zhang2025swe}, math \citep{lightman2023let} or reasoning \citep{lin2025zebralogic} exist, these tasks are still challenging for full LLMs \citep{yang2025qwen3, liu2024deepseek}.
Moreover, existing study shows that model compression for complex tasks often requires additional task-specific fine-tuning \citep{sarkar2024revisiting, chen2022task}, which is beyond the scope of this work and we consider it a promising direction of future work.

\subsection{Effectiveness evaluation} \label{effect exp}

\begin{table}[t]
\footnotesize
\centering
\caption{
    Zero shot performance with 100 calibration samples from Zyda2 dataset. 
    Best results are in \textbf{bold}, and the second best are \underline{underlined}. 
    \colorbox{green!25}{Green cells} indicate results no less than original models.}
\label{combined-tables}
\subfloat[OLMoE]{
\label{olmoe-zyda2-100}
\resizebox{\textwidth}{!}{
    \begin{tabular}{@{}ccc*{10}{w{c}{0.95cm}}@{}}
    \toprule
    Pruning ratio          & Model/Method  & Arc-c & Arc-e & BoolQ & COPA  & MMLU  & OBQA  & PIQA  & RTE   & Winogrande & Avg. \\ \midrule
    0\%                   & OLMoE   & 49.23 & 76.89 & 70.09 & 85.00 & 53.54 & 44.40 & 79.76 & 71.84 & 68.90      & 66.63   \\ \midrule
    \multirow{5}{*}{25\%} & Angular & 32.76 & 61.91 & 61.71 & 74.00 & 23.13 & 37.60 & 71.65 & 53.07 & 55.09      & 52.33   \\
                          & FLAP    & \underline{40.53} & \textbf{67.55} & \underline{62.69} & \underline{78.00} & \textbf{41.16} & \underline{37.80} & \underline{74.81} & \underline{61.37} & 60.93      & \underline{58.32}   \\
                          & MC-SMoE & 35.67 & 54.92 & 63.49 & 73.00 & 29.04 & 30.60 & 67.19 & 55.23 & \underline{65.75}      & 52.77   \\
                          & RS      & 25.85 & 43.01 & 59.08 & 74.00 & 29.63 & 36.20 & 66.16 & 56.68 & 59.98      & 50.07   \\
                          & MoNE (Ours)    & \textbf{42.32} & \underline{64.81} & \textbf{67.19} & \cellcolor{green!25}\textbf{85.00} & \underline{40.13} & \textbf{40.80} & \textbf{78.07} & \textbf{64.62} & \textbf{66.46}      & \textbf{61.04}     \\ \bottomrule
    \end{tabular}
}
}

\subfloat[Moonlight]{
\label{moonlight-zyda2-100}
\resizebox{\textwidth}{!}{
    \begin{tabular}{@{}ccc*{10}{w{c}{0.95cm}}@{}}
    \toprule
    Pruning ratio          & Model/Method & Arc-c & Arc-e & BoolQ & COPA  & MMLU  & OBQA  & PIQA  & RTE   & Winogrande & Avg. \\ \midrule
    0\%                   & Moonlight    & 58.28 & 82.49 & 80.40 & 92.00 & 67.30 & 45.60 & 81.12 & 65.70 & 71.11      & 71.56   \\ \midrule
    \multirow{5}{*}{25\%} & Angular      & 39.76 & 52.69 & 38.90 & 79.00 & 42.57 & 32.20 & 68.50 & \underline{61.01} & 62.04      & 52.96   \\
                          & FLAP         & 48.55 & 76.01 & 75.93 & \textbf{90.00} & \textbf{55.84} & 42.20 & 77.97 & \textbf{64.26} & 68.19      & 66.55   \\
                          & MC-SMoE      & 47.61 & 73.15 & \underline{78.72} & \underline{89.00} & 46.11 & 43.60 & 80.36 & 56.32 & \cellcolor{green!25}71.43      & 65.14   \\
                          & RS           & \underline{55.80} & \textbf{80.64} & 78.69 & \textbf{90.00} & 46.73 & \cellcolor{green!25}\underline{46.40} & \textbf{81.01} & 58.84 & \cellcolor{green!25}\textbf{72.30}      & \underline{67.82}   \\
                          & MoNE (Ours)         & \textbf{55.89} & \underline{80.60} & \textbf{79.57} & \textbf{90.00} & \underline{55.23} & \cellcolor{green!25}\textbf{46.80} & \underline{80.85} & \underline{61.01} & \cellcolor{green!25}\underline{71.98}      & \textbf{69.10}  \\ \bottomrule
    \end{tabular}
}
} 

\subfloat[Deepseek-V2-Lite]{
\label{dsv2lite-zyda2-100}
\resizebox{\textwidth}{!}{
    \begin{tabular}{@{}ccc*{10}{w{c}{0.95cm}}@{}}
    \toprule
    Pruning ratio          & Model/Method     & Arc-c & Arc-e & BoolQ & COPA  & MMLU  & OBQA  & PIQA  & RTE   & Winogrande & Avg. \\ \midrule
    0\%                   & Deepseek-V2-Lite & 48.72 & 76.18 & 79.88 & 88.00 & 54.96 & 43.60 & 80.25 & 61.37 & 71.51      & 67.16   \\ \midrule
    \multirow{5}{*}{25\%} & Angular          & 32.00 & 53.28 & 64.92 & 75.00 & 26.95 & 34.00 & 71.33 & 58.84 & 61.01      & 53.04   \\
                          & FLAP             & 43.69 & 71.46 & \underline{75.26} & \underline{84.00} & 47.28 & 41.40 & 78.18 & \cellcolor{green!25}\textbf{62.82} & 67.72      & 63.53   \\
                          & MC-SMoE          & 36.69 & 60.77 & 71.31 & \underline{84.00} & 42.22 & 36.60 & 75.57 & 58.48 & 68.67      & 59.37   \\
                          & RS               & \cellcolor{green!25}\textbf{49.32} & \underline{74.41} & 69.39 & \cellcolor{green!25}\textbf{90.00} & \textbf{50.35} & \cellcolor{green!25}\textbf{43.80} & \textbf{80.14} & \cellcolor{green!25}\underline{62.09} & \underline{70.24}      & \underline{65.53}   \\
                          & MoNE (Ours)             & \underline{46.67} & \textbf{74.62} & \textbf{78.47} & \cellcolor{green!25}\textbf{90.00} & \underline{49.05} & \underline{43.00} & \underline{79.76} & \cellcolor{green!25}\underline{62.09} & \textbf{71.43}      & \textbf{66.12}  \\ \bottomrule
    \end{tabular}
}
} 

\subfloat[Qwen2-57B-A14B]{
\label{qwen2-57b-zyda2-100}
\resizebox{\textwidth}{!}{
    \begin{tabular}{@{}ccc*{10}{w{c}{0.95cm}}@{}}
    \toprule
    Pruning ratio          & Model/Method & Arc-c & Arc-e & BoolQ & COPA  & MMLU  & OBQA  & PIQA  & RTE   & Winogrande & Avg. \\ \midrule
    0\%                   & Qwen2-57B-A14B    & 49.66 & 69.44 & 86.45 & 93.00 & 74.06 & 44.20 & 81.23 & 74.73 & 74.27      & 71.89   \\ \midrule
    \multirow{5}{*}{25\%} & Angular      & 29.44 & 54.17 & 59.51 & 70.00 & 23.92 & 32.80 & 70.02 & 54.87 & 49.57 & 49.37   \\
                          & FLAP         & \cellcolor{green!25}\textbf{50.00} & \cellcolor{green!25}\textbf{72.85} & \cellcolor{green!25}\textbf{86.91} & \underline{91.00} & 65.02 & \cellcolor{green!25}\textbf{45.40} & \underline{81.12} & \cellcolor{green!25}\textbf{77.62} & \cellcolor{green!25}\textbf{76.09} & \textbf{71.78}   \\
                          & MC-SMoE      & 46.67 & 66.25 & 86.45 & 88.00 & 69.46 & 43.40 & 80.20 & \cellcolor{green!25}74.73 & \cellcolor{green!25}\underline{75.14} & 70.03   \\
                          & RS           & 49.15 & \cellcolor{green!25}\underline{69.78} & 84.77 & 87.00 & \underline{70.99} & \cellcolor{green!25}44.80 & \cellcolor{green!25}\textbf{81.34} & 74.37 & 74.19 & 70.71   \\
                          & MoNE (Ours)         & \cellcolor{green!25}\underline{49.66} & 68.73 & \cellcolor{green!25}\underline{86.88} & \cellcolor{green!25}\textbf{94.00} & \textbf{71.64} & \cellcolor{green!25}\underline{45.20} & 81.07 & \cellcolor{green!25}\underline{75.45} & 73.09 & \underline{71.75}  \\ \bottomrule
    \end{tabular}
}
} 

\subfloat[Qwen3-30B-A3B]{
\label{qwen3-30b-zyda2-100}
\resizebox{\textwidth}{!}{
    \begin{tabular}{@{}ccc*{10}{w{c}{0.95cm}}@{}}
    \toprule
    Pruning ratio          & Model/Method & Arc-c & Arc-e & BoolQ & COPA  & MMLU  & OBQA  & PIQA  & RTE   & Winogrande & Avg. \\ \midrule
    0\%                   & Qwen3-30B-A3B    & 55.89 & 79.42 & 88.69 & 84.00 & 77.82 & 44.80 & 80.30 & 82.31 & 70.64 & 73.76   \\ \midrule
    \multirow{5}{*}{25\%} & Angular      & 44.62 & 68.60 & 80.52 & 77.00 & 59.75 & 40.40 & 75.30 & 70.40 & 62.51 & 64.34   \\
                          & FLAP         & 50.85 & 76.68 & 85.72 & \cellcolor{green!25}\textbf{85.00} & 69.43 & 42.80 & 77.31 & \textbf{81.95} & 70.17 & 71.10   \\
                          & MC-SMoE      & 52.73 & 76.98 & \cellcolor{green!25}\underline{88.75} & \underline{83.00} & 72.25 & \textbf{44.40} & \underline{79.71} & \underline{80.87} & \underline{70.40} & 72.12   \\
                          & RS           & \underline{53.75} & \underline{78.32} & 88.53 & \cellcolor{green!25}\textbf{85.00} & \textbf{74.60} & \underline{43.00} & \textbf{79.92} & 79.78 & 69.61 & \underline{72.50}   \\
                          & MoNE (Ours)         & \cellcolor{green!25}\textbf{56.14} & \textbf{79.17} & \cellcolor{green!25}\textbf{89.11} & \cellcolor{green!25}\textbf{85.00} & \underline{74.04} & \underline{43.00} & 79.27 & 77.98 & \textbf{70.48} & \textbf{72.69}  \\ \bottomrule
    \end{tabular}
}
}
\end{table}

\begin{table}[t]
\footnotesize
\centering
\caption{
    Maximum pruning ratios with 1\% accuracy loss after MoNE pruning using 100 calibration samples from Zyda-2 dataset.}
\label{max-pruning-ratio}
\begin{tabular}{lccc}
\toprule
Model & Max. pruning ratio & Avg. perf before pruning & Avg. perf after pruning \\ \midrule
OLMoE	 & 16\%	 & 66.63 & 66.00   \\
Moonlight	 & 16\%	 & 71.56	 & 70.59   \\
Deepseek-V2-Lite	 & 20\%	 & 67.16	 & 66.31 \\
Qwen2-57B-A14B & 25\%	 & 71.89 & 71.75   \\
Qwen3-30B-A3B & 24\% & 73.76 & 73.61     \\ \bottomrule
\end{tabular}
\end{table}

This section validates the effectiveness of MoNE by comparing the zero shot performance of 25\% pruned models with 100 calibration samples from the Zyda2 dataset.
The results are presented in Table \ref{combined-tables} and Table \ref{max-pruning-ratio}.
Results under 50\% pruning ratio are extended to Table \ref{zyda2-100} in Appendix \ref{robust detail}.

Table \ref{combined-tables} indicates that \textbf{MoNE consistently outperforms baseline methods} in terms of the average accuracy on the nine tasks. 
In particular, it shows average accuracy improvement as large as \textbf{2.72} for the pruned OLMoE compared to baseline methods, and it incurs accuracy drop as small as only \textbf{0.14} for the pruned Qwen2-57B-A14B.
Furthermore, MoNE-pruned models can achieve \textbf{either the best or the second best result for individual tasks under most settings}.

An interesting observation is that all the three expert pruning methods, MC-SMoE, RS and MoNE can achieve results on par with or even better than the original models on certain tasks. 
The specific examples are shown in green background in Table \ref{combined-tables}.
All these results indicate that there is indeed redundancy existing in the expert level for the examined MoE models, and expert pruning can rule out such redundancy to achieve even better results on these tasks.
Among the three expert pruning methods, MoNE consistently surpasses the two baseline methods, demonstrating its strong capability.

Table \ref{max-pruning-ratio} evaluates the maximum pruning ratios that maintain accuracy loss within 1\% across different MoE models. 
As shown in Table \ref{max-pruning-ratio}, larger and more powerful models such as Qwen2-57B-A14B and Qwen3-30B-A3B tolerate more aggressive pruning (25\% and 24\%, respectively) compared to smaller models like OLMoE, Moonlight, and Deepseek-V2-Lite (16–20\%). 
This finding highlights the scalability of proposed MoNE method for large model scales.
In addition, combining the observations from Table \ref{combined-tables} and Table \ref{max-pruning-ratio}, we argue that larger MoE models may have stronger capability in language modeling but also contain increasing parameter redundancy at the expert level, and MoNE can efficiently eliminates such redundancy at minimal performance degradation.

\subsection{Robustness evaluation} \label{calib size}
\begin{figure}[t]
    \vspace{-2mm}
    \centering
    \begin{subfigure}[t]{\textwidth}
        \includegraphics[width=\textwidth]{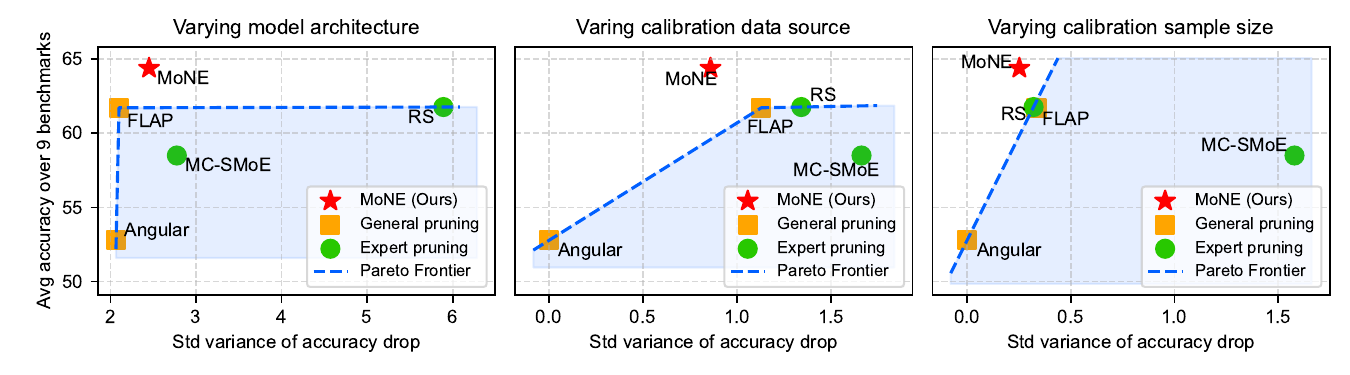}
        \vspace{-7mm}
        \caption{25\% pruning ratio.}
        \label{fig:robust_p25}
    \end{subfigure}
    
    \begin{subfigure}[t]{\textwidth}
        \includegraphics[width=\textwidth]{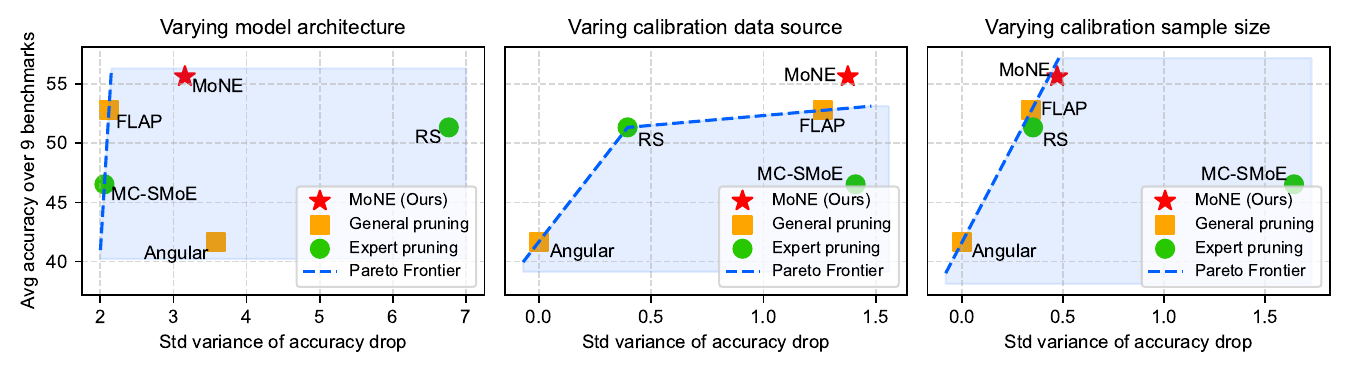}
        \vspace{-7mm}
        \caption{50\% pruning ratio.}
        \label{fig:robust_p50}
    \end{subfigure}
    \vspace{-2mm}
    \caption{
        Average accuracy versus accuracy drop variance.
        MoNE advances the Pareto frontier across varying model architectures, calibration data sources and calibration sample sizes.
    }
    \label{fig:robust}
    \vspace{-4mm}
\end{figure}

This section evaluates the robustness of MoNE across three key dimensions: model architecture, calibration data source, and calibration sample size. 
Due to prohibitive compute for exhaustive experiments on large models, we tested three models: OLMoE, Moonlight and Deepseek-V2-Lite using 100, 500 and 1000 calibration samples on C4 and Zyda-2 dataset separately.
For each dimension, we vary one factor while averaging results over the other two, measuring both average accuracy and the standard deviation of accuracy drop. 
The results are visualized in Figure \ref{fig:robust}, with detailed scores provided in Appendix \ref{robust detail}.
As shown in Figure \ref{fig:robust_p25}, \textbf{MoNE advances the Pareto frontier across all three dimensions at the 25\% pruning ratio}, demonstrating superior robustness and effectiveness compared to existing structured pruning methods.
At the 50\% pruning ratio (Figure \ref{fig:robust_p50}), MoNE exhibits slightly higher variance under varying model architectures and calibration sample sizes. 
Nevertheless, it remains the most effective method, \textbf{outperforming baseline methods by a significant margin of 2.85}.

\subsection{Ablation study}
\begin{figure}[t]
    \vspace{-2mm}
    \centering
    \includegraphics[width=0.9\textwidth]{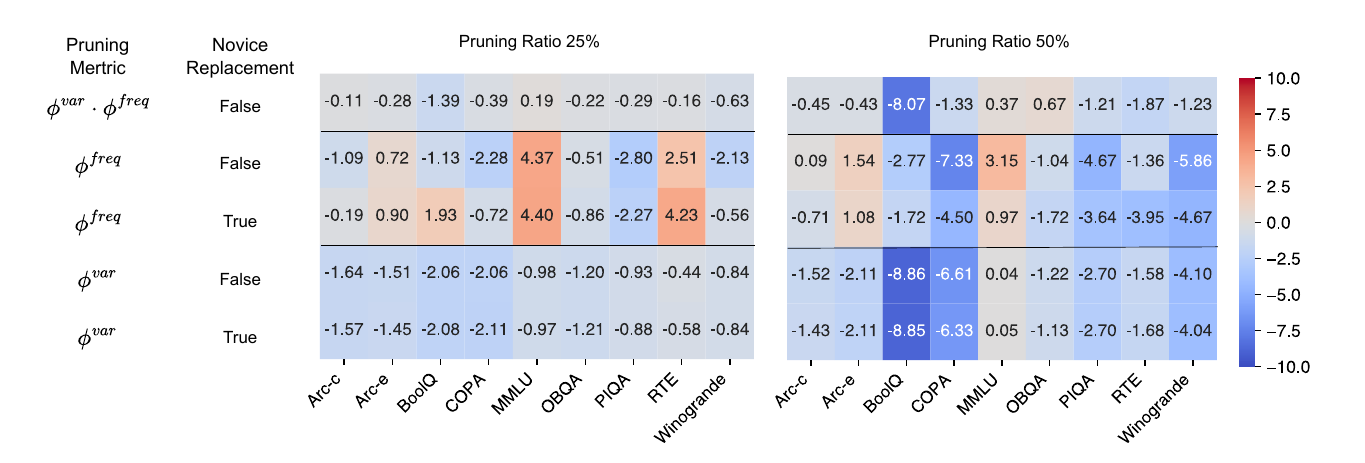}
    \vspace{-6mm}
    \caption{
        Ablation study on expert access frequency, output variance and novice replacement. 
        Numbers are the difference to the proposed MoNE. 
        The detailed result is provided in Appendix \ref{ablation study detail}.
    }
    \label{ablation}
    \vspace{-3mm}
\end{figure}

This section presents the ablation study to evaluate the effects of the two redundancy metrics and the impact of novice replacement across the downstream tasks. Figure \ref{ablation} displays the average accuracy drop relative to our proposed methods, with lower values indicating greater degradation. Results are averaged over the three evaluation dimensions to provide a robust assessment. 
We observe that integrating the fused expert redundancy score with novice replacement yields better performance, particularly under higher pruning ratios. This indicates that our approach is especially effective in preserving model quality when pruning is more aggressive.
Notably, for tasks such as BoolQ, COPA, and PIQA, our proposed method outperforms the ablation baselines by a large margin—achieving accuracy gains of up to 8.85. However, for MMLU, pruning based solely on frequency appears to offer a slight advantage, suggesting that frequently activated experts may play a more critical role in domain-specific reasoning tasks.

\begin{table}[t]
\centering
\footnotesize
\caption{
    Zero shot performance of the 25\% pruned OLMoE after continued pretraining with 2B tokens from OLMoE-mix-0924.
    Best results are in \textbf{bold}, and the second best are \underline{underlined}. 
}
\label{continue pretrain}
\resizebox{\textwidth}{!}{
    \begin{tabular}{@{}ccccccccccc@{}}
    \toprule
    Model/Method & Arc-c & Arc-e & BoolQ & COPA  & MMLU  & OBQA  & PIQA  & RTE   & Winogrande & Average \\ \midrule
    OLMoE        & 49.23 & 76.89 & 70.09 & 85.00 & 53.54 & 44.40 & 79.76 & 71.84 & 68.90      & 66.63   \\ \midrule
    Angular      & 38.82 & 64.69 & 63.52 & 82.00 & 25.42 & 39.80 & 76.50 & 51.62 & 59.04      & 55.71   \\
    FLAP         & 42.24 & 69.07 & 69.51 & 80.00 & \textbf{45.56} & \underline{40.40} & 77.42 & 50.18 & 63.54      & 59.77   \\
    MC-SMoE      & 42.75 & 70.41 & 69.76 & 80.00 & \underline{44.13} & 37.60 & 75.79 & 66.43 & 64.96      & 61.31   \\
    RS           & \underline{44.97} & \underline{72.94} & \underline{70.73} & \underline{85.00} & 43.28 & \textbf{43.00} & \underline{78.67} & \textbf{72.20} & \underline{65.98}      & \underline{64.09}   \\
    MoNE (Ours)         & \textbf{47.35} & \textbf{74.33} & \textbf{71.56} & \textbf{87.00} & 43.30 & \underline{40.40} & \textbf{78.89} & \underline{67.51} & \textbf{67.25}      & \textbf{64.18}   \\ \bottomrule
    \end{tabular}
}
\end{table}

\subsection{Accuracy recovery with continued pretraining}

To evaluate performance recovery capabilities, we conducted continued pretraining on the 25\% compressed OLMoE model pruned by 100 Zyda2 samples, as only this model releases its pretraining dataset, OLMoE-mix-0924\footnote{\url{https://huggingface.co/datasets/allenai/OLMoE-mix-0924}}.
The sequence length was set to 4096 and the global token size per step was 4M.
Each pruned model was trained with 2B tokens, i.e., 512 steps, and the peak and minimum learning rate (lr) were 5e-5 and 5e-6, respectively.
We employed the cosine lr scheduler with 50 warm up steps.
Other hyperparameters were the same as the original configuration for OLMoE \citep{muennighoff2024olmoe}.
All the experiments could run on a single H20 GPU, but we accelerated the training with 16 H20 GPUs.

The results are summarized in Table \ref{continue pretrain}.
This table shows that MoNE achieves the average accuracy closest to the original model with only 2B tokens from a pretraining dataset, demonstrating the promising capability of the MoNE computation framework.
Besides, MC-SMoE and RS reclaim 8.54 and 14.02 average accuracy, indicating that expert pruning is not only effective to eliminate redundancy, but also relatively easier to recover performance with continued pretraining.

%% file: sections/conclusion.tex
\section{Conclusion}
In this paper, we propose MoNE, a novel expert pruning method that replaces redundant experts with lightweight novices to compress MoE models. 
MoNE evaluates expert redundancy based on expert access frequency and output variance in each model layer, pruning experts with low usage and stable outputs while replacing them with novices that provide unbiased output estimates. Extensive experiments across different MoE architectures, calibration data sources, and sample sizes demonstrate that MoNE outperforms existing structured pruning methods by maintaining higher zero-shot performance across nine downstream tasks.

%% file: sections/appendix.tex
\newpage 
\section*{The Use of Large Language Models}
All writing, visualizations, and experiments are \textbf{completed by the authors}. LLMs (e.g., Claude) are used \textbf{solely to refine the writing}.

We organize our appendix as follows:
\begin{itemize}
    \item Section \ref{subsec:specialized_tasks} shows the evaluation of MoNE on domain-specific tasks.
    \item Section \ref{pruning_metric_ablation} presents ablation study on the pruning metric.
    \item Section \ref{robust detail} provides detailed scores for various models, calibration datasets and calibration sample sizes.
    \item Section \ref{redundant experts} visualizes expert access frequency and output variance for more tasks and models.
    \item Section \ref{ablation study detail} presents the detailed scores for ablation study.
    \item Section \ref{deployment} shows the inference latency and memory overhead benefits for the proposed MoNE.
\end{itemize}

\newpage
\section{Specialized task evaluation}
\label{subsec:specialized_tasks}

\begin{table}[!t]
\centering
\caption{Base and Instruct OLMoE models on Math and GSM8K.}
\label{tab:olmoe_specialized}
\begin{tabular}{lcc}
\toprule
\textbf{Model} & \textbf{GSM8K} & \textbf{Math} \\
\midrule
\multicolumn{3}{l}{\textit{No calibration}} \\
OLMoE & 52.77 & 15.78 \\
OLMoE-Instruct & 67.63 & 18.64 \\
\midrule
\multicolumn{3}{l}{\textit{Calibration: Zyda2}} \\
OLMoE & 3.34 & 1.94 \\
OLMoE-Instruct & 7.96 & 2.54 \\
\midrule
\multicolumn{3}{l}{\textit{Calibration: Task-specific}} \\
OLMoE-Instruct (Math) & 65.28 & 18.22 \\
OLMoE-Instruct (GSM8K) & 67.48 & 17.64 \\
\bottomrule
\end{tabular}
\end{table}

\begin{table}[!t]
\centering
\caption{Base and Instruct Moonlight models on Math and GSM8K.}
\label{tab:moonlight_specialized}
\begin{tabular}{lcc}
\toprule
\textbf{Model} & \textbf{GSM8K} & \textbf{Math} \\
\midrule
\multicolumn{3}{l}{\textit{No calibration}} \\
Moonlight & 74.22 & 42.32 \\
Moonlight-Instruct & 77.03 & 39.26 \\
\midrule
\multicolumn{3}{l}{\textit{Calibration: Zyda2}} \\
Moonlight & 46.40 & 3.32 \\
Moonlight-Instruct & 51.86 & 6.64 \\
\midrule
\multicolumn{3}{l}{\textit{Calibration: Task-specific}} \\
Moonlight-Instruct (Math) & 75.28 & 38.56 \\
Moonlight-Instruct (GSM8K) & 76.72 & 35.84 \\
\bottomrule
\end{tabular}
\end{table}

\begin{table}[!t]
\centering
\caption{Base and Instruct DeepSeek-V2-Lite models on Math and GSM8K.}
\label{tab:deepseek_specialized}
\begin{tabular}{lcc}
\toprule
\textbf{Model} & \textbf{GSM8K} & \textbf{Math} \\
\midrule
\multicolumn{3}{l}{\textit{No calibration}} \\
DeepSeek-V2-Lite & 38.82 & 16.54 \\
DeepSeek-V2-Lite-Chat & 66.49 & 18.56 \\
\midrule
\multicolumn{3}{l}{\textit{Calibration: Zyda2}} \\
DeepSeek-V2-Lite & 25.40 & 6.80 \\
DeepSeek-V2-Lite-Chat & 37.07 & 4.76 \\
\midrule
\multicolumn{3}{l}{\textit{Calibration: Task-specific}} \\
DeepSeek-V2-Lite-Chat (Math) & 64.44 & 20.76 \\
DeepSeek-V2-Lite-Chat (GSM8K) & 63.76 & 19.70 \\
\bottomrule
\end{tabular}
\end{table}

\begin{table}[!t]
\centering
\caption{Base and Instruct Qwen2-57B-A14B models on Math and GSM8K.}
\label{tab:qwen2_specialized}
\begin{tabular}{lcc}
\toprule
\textbf{Model} & \textbf{GSM8K} & \textbf{Math} \\
\midrule
\multicolumn{3}{l}{\textit{No calibration}} \\
Qwen2-57B-A14B & 79.68 & 40.50 \\
Qwen2-57B-A14B-Instruct & 69.90 & 31.30 \\
\midrule
\multicolumn{3}{l}{\textit{Calibration: Zyda2}} \\
Qwen2-57B-A14B & 74.07 & 31.28 \\
Qwen2-57B-A14B-Instruct & 64.90 & 22.84 \\
\midrule
\multicolumn{3}{l}{\textit{Calibration: Task-specific}} \\
Qwen2-57B-A14B-Instruct (Math) & 58.45 & 29.56 \\
Qwen2-57B-A14B-Instruct (GSM8K) & 52.92 & 29.62 \\
\bottomrule
\end{tabular}
\end{table}

\begin{table}[!t]
\centering
\caption{Base and Instruct Qwen3-30B-A3B models on Math and GSM8K.}
\label{tab:qwen3_specialized}
\begin{tabular}{lcc}
\toprule
\textbf{Model} & \textbf{GSM8K} & \textbf{Math} \\
\midrule
\multicolumn{3}{l}{\textit{No calibration}} \\
Qwen3-30B-A3B & 85.37 & 50.48 \\
Qwen3-30B-A3B-Instruct & 90.90 & 46.94 \\
\midrule
\multicolumn{3}{l}{\textit{Calibration: Zyda2}} \\
Qwen3-30B-A3B & 71.72 & 9.44 \\
Qwen3-30B-A3B-Instruct & 78.01 & 9.88 \\
\midrule
\multicolumn{3}{l}{\textit{Calibration: Task-specific}} \\
Qwen3-30B-A3B-Instruct (Math) & 89.76 & 48.52 \\
Qwen3-30B-A3B-Instruct (GSM8K) & 90.30 & 51.70 \\
\bottomrule
\end{tabular}
\end{table}

We extended MoNE to two specialized tasks: Math and GSM8K. We conducted experiments on both base models and instruct models. To calibrate the instruct models, we adopted the first 100 samples from the training dataset of each task. The results are reported in Tables~\ref{tab:olmoe_specialized}--\ref{tab:qwen3_specialized}.

Our experiments on specialized tasks reveal two important findings regarding the effectiveness of different calibration strategies. After pruning with pretraining data (Zyda2), larger base models preserve significantly more accuracy, which is consistent with our observations from general tasks in Section \ref{effect exp}. However, pretraining data cannot accurately capture the distribution of specialized tasks, leading to accuracy drops of up to 49\% for the smallest model, OLMoE. This demonstrates that while model scale provides some resilience to pruning, the mismatch between pretraining data and task-specific distributions poses substantial challenges for maintaining performance on specialized tasks.

In contrast, MoNE can effectively preserve performance by pruning instruct models with training data from the specialized tasks themselves. For example, OLMoE-Instruct incurs only minimal accuracy drops of at most 1\% for both tasks when calibrated with task-specific data. This result is not surprising, as state-of-the-art models also rely on specialized fine-tuning approaches such as supervised fine-tuning and reinforcement learning with domain-specific data to enhance their capability in these tasks \citep{muennighoff2024olmoe, liu2025muon, team2024qwen2, yang2025qwen3}. The effectiveness of task-specific calibration suggests that the distribution alignment between calibration data and target tasks plays a critical role in determining post-pruning performance.

In summary, we acknowledge that calibration via pretraining data is insufficient to preserve model capability for specialized tasks, and task-specific calibration can effectively mitigate this issue. A promising future direction for model compression is to develop methods that bridge the gap between these two calibration approaches, potentially through hybrid calibration strategies or adaptive data selection techniques that better capture task-specific distributions while maintaining broad coverage.

\section{Pruning metric ablation study}
\label{pruning_metric_ablation}

We conducted an extended ablation study on the redundancy score to evaluate the sensitivity of MoNE to different scoring formulations. Following the same evaluation configuration as in Section \ref{effect exp}, we compared MoNE against five variants: normalized scores, log-sum aggregation, and weighted sum with varying emphasis on output variance (25\%, 50\%, and 75\%). The results are reported in Table~\ref{tab:ablation_redundancy}.

\begin{table}[htbp]
\centering
\caption{Ablation study on redundancy scoring methods across different models. All values represent average performance scores.}
\label{tab:ablation_redundancy}
\resizebox{\textwidth}{!}{
\begin{tabular}{lcccccc}
\toprule
\textbf{Model} & \textbf{MoNE} & \textbf{Normalized} & \textbf{Log-sum} & \textbf{Weighted 25\%} & \textbf{Weighted 50\%} & \textbf{Weighted 75\%} \\
\midrule
OLMoE & 61.04 & 61.39 & 61.23 & 57.07 & 57.07 & 57.07 \\
Moonlight & 69.10 & 68.58 & 69.05 & 68.95 & 69.03 & 69.07 \\
DeepSeek-V2-Lite & 66.12 & 65.70 & 66.02 & 65.63 & 65.66 & 66.07 \\
Qwen2-57B-A14B & 71.75 & 71.73 & 71.75 & 72.55 & 72.57 & 72.53 \\
Qwen3-30B-A3B & 72.69 & 73.62 & 73.66 & 71.96 & 72.29 & 72.48 \\
\bottomrule
\end{tabular}
}
\end{table}

The results indicate that the pruning score exhibits minimal scale sensitivity across different formulations. In particular, the log-sum aggregation shows nearly identical results to MoNE, which is expected since $\log(\theta^{var}) + \log(\theta^{freq}) = \log(\theta^{var} \times \theta^{freq})$ does not affect the partial ordering during expert redundancy ranking. This mathematical equivalence ensures that both formulations produce the same pruning decisions despite their different representations.

We did not include learnable weights in this ablation study, as training five models with learnable parameters would be prohibitively expensive and time-consuming. However, the weighted sum results reveal model-dependent behavior, where only Qwen2-57B-A14B achieves minor improvements of approximately 0.8\% with weighted scoring. In contrast, smaller models like OLMoE incur severe performance drops of around 4\% under the weighted sum formulation. This suggests that while larger models may benefit from adjusting the relative importance of variance and frequency components, smaller models are more sensitive to such modifications and perform better with the balanced geometric mean approach used in MoNE.

\section{More detailed results} \label{robust detail}

\begin{table}
\footnotesize
\centering
\caption{
    Zero shot performance with 100 calibration samples from Zyda2 dataset.
    Best results are in \textbf{bold}, and the second best are \underline{underlined}. 
}
\label{zyda2-100}
\subfloat[OLMoE]{
\resizebox{\textwidth}{!}{
\begin{tabular}{@{}ccc*{10}{w{c}{0.95cm}}@{}}
\toprule
Pruning ratio         & Model/Method & Arc-c          & Arc-e          & BoolQ          & COPA           & MMLU           & OBQA           & PIQA           & RTE            & Winogrande     & Average        \\ \midrule
0\%                   & OLMoE        & 49.23          & 76.89          & 70.09          & 85.00          & 53.54          & 44.40          & 79.76          & 71.84          & 68.90          & 66.63          \\ \midrule
\multirow{5}{*}{50\%} & Angular      & 27.22          & 37.50          & 53.91          & 62.00          & 23.96          & 26.60          & 58.27          & 52.35          & 51.85          & 43.74          \\
                      & FLAP         & \textbf{29.18} & \textbf{54.92} & \underline{62.17} & \underline{68.00} & \textbf{30.51} & \underline{29.60} & \textbf{67.57} & \underline{55.23} & \underline{56.27} & \underline{50.39} \\
                      & MC-SMoE      & 24.49          & 31.44          & 59.33          & 67.00          & 23.01          & 26.00          & 53.92          & 53.43          & 53.12          & 43.53          \\
                      & RS           & 21.50          & 28.62          & 39.45          & 61.00          & 23.27          & 26.00          & 52.34          & 51.99          & 51.85          & 39.56          \\
                      & MoNE (Ours)  & \underline{28.16} & \underline{40.24} & \textbf{63.12} & \textbf{78.00} & \underline{25.21} & \textbf{32.40} & \underline{63.33} & \textbf{60.65} & \textbf{63.54} & \textbf{50.52} \\ \bottomrule
\end{tabular}
}
}

\subfloat[Moonlight]{
\resizebox{\textwidth}{!}{
    \begin{tabular}{@{}ccc*{10}{w{c}{0.95cm}}@{}}
\toprule
Pruning ratio         & Model/Method & Arc-c          & Arc-e          & BoolQ          & COPA           & MMLU           & OBQA           & PIQA           & RTE            & Winogrande     & Average        \\ \midrule
0\%                   & Moonlight    & 58.28          & 82.49          & 80.40          & 92.00          & 67.30          & 45.60          & 81.12          & 65.70          & 71.11          & 71.56          \\ \midrule
\multirow{5}{*}{50\%} & Angular      & 27.90          & 28.54          & 48.01          & 49.00          & \underline{25.67} & 28.80          & 52.56          & 51.99          & 47.75          & 40.02          \\
                      & FLAP         & 33.87          & \underline{61.36}          & 63.30          & 75.00          & \textbf{36.80} & 36.00          & 69.37          & 57.04          & 62.12          & 54.98          \\
                      & MC-SMoE      & 29.52          & 47.94          & 59.54          & 79.00          & 23.94          & 31.40          & 67.30          & 57.04          & 60.46          & 50.68          \\
                      & RS           & \underline{37.80} & 58.42          & \underline{70.86} & \underline{89.00} & 23.27          & \underline{38.00} & \underline{78.18} & \underline{57.76} & \textbf{70.80} & \underline{58.23} \\
                      & MoNE (Ours)  & \textbf{43.09} & \textbf{70.03} & \textbf{76.12} & \textbf{90.00} & 23.57          & \textbf{40.80} & \textbf{78.78} & \textbf{58.84} & \underline{70.17} & \textbf{61.27} \\ \bottomrule
\end{tabular}
}
} 

\subfloat[Deepseek-V2-Lite]{
\resizebox{\textwidth}{!}{
    \begin{tabular}{@{}ccc*{10}{w{c}{0.95cm}}@{}}
\toprule
Pruning ratio         & Model/Method     & Arc-c          & Arc-e          & BoolQ          & COPA           & MMLU           & OBQA           & PIQA           & RTE            & Winogrande     & Average        \\ \midrule
0\%                   & Deepseek-V2-Lite & 48.72          & 76.18          & 79.88          & 88.00          & 54.96          & 43.60          & 80.25          & 61.37          & 71.51          & 67.16          \\ \midrule
\multirow{5}{*}{50\%} & Angular          & 24.06          & 32.79          & 40.40          & 61.00          & 23.22          & 26.80          & 56.42          & \underline{57.76} & 49.09          & 41.28          \\
                      & FLAP             & 35.24          & \underline{60.31} & \underline{69.66} & 79.00          & \underline{36.13} & 35.20          & 74.76          & 56.68          & \underline{64.09} & \underline{56.79} \\
                      & MC-SMoE          & 24.57          & 35.82          & 56.36          & 67.00          & 28.29          & 27.40          & 61.37          & 52.35          & 53.91          & 45.23          \\
                      & RS               & \underline{36.01} & 57.45          & 57.98          & \textbf{89.00} & 24.91          & \textbf{40.80} & \textbf{78.02} & 54.15          & 62.75          & 55.67          \\
                      & MoNE (Ours)      & \textbf{37.20} & \textbf{67.17} & \textbf{73.39} & \underline{84.00} & \textbf{42.30} & \underline{36.80} & \underline{75.30} & \textbf{59.57} & \textbf{67.88} & \textbf{60.40} \\ \bottomrule
\end{tabular}
}
} 

\subfloat[Qwen2-57B-A14B]{
\resizebox{\textwidth}{!}{
    \begin{tabular}{@{}ccc*{10}{w{c}{0.95cm}}@{}}
\toprule
Pruning ratio         & Model/Method   & Arc-c          & Arc-e          & BoolQ          & COPA           & MMLU           & OBQA           & PIQA           & RTE            & Winogrande     & Avg.           \\ \midrule
0\%                   & Qwen2-57B-A14B & 49.66          & 69.44          & 86.45          & 93.00          & 74.06          & 44.20          & 81.23          & 74.73          & 74.27          & 71.89          \\ \midrule
\multirow{5}{*}{50\%} & Angular        & 25.17          & 31.19          & 45.08          & 54.00          & 23.74          & 29.80          & 54.62          & 53.07          & 50.28          & 40.77          \\
                      & FLAP           & 36.35          & 61.95          & 73.43          & 80.00          & 46.50          & 38.20          & 74.59          & 68.95          & \underline{68.59} & 60.95          \\
                      & MC-SMoE        & \underline{41.64} & \underline{64.73} & \underline{79.91} & \textbf{87.00} & \underline{59.65} & \underline{38.80} & \underline{77.53} & \textbf{71.84} & 68.27          & \underline{65.49} \\
                      & RS             & 20.90          & 29.38          & 50.83          & 57.00          & 23.39          & 25.40          & 52.88          & 51.26          & 49.25          & 40.03          \\
                      & MoNE (Ours)    & \textbf{45.43} & \textbf{65.22} & \textbf{80.99} & \underline{85.00} & \textbf{62.00} & \textbf{42.00} & \textbf{78.41} & \underline{70.90} & \textbf{69.51} & \textbf{66.61} \\ \bottomrule
\end{tabular}
}
} 

\subfloat[Qwen3-30B-A3B]{
\resizebox{\textwidth}{!}{
    \begin{tabular}{@{}ccc*{10}{w{c}{0.95cm}}@{}}
\toprule
Pruning ratio         & Model/Method  & Arc-c          & Arc-e          & BoolQ          & COPA           & MMLU           & OBQA           & PIQA           & RTE            & Winogrande     & Avg.           \\ \midrule
0\%                   & Qwen3-30B-A3B & 55.89          & 79.42          & 88.69          & 84.00          & 77.82          & 44.80          & 80.30          & 82.31          & 70.64          & 73.76          \\ \midrule
\multirow{5}{*}{50\%} & Angular       & 26.37          & 39.48          & 50.86          & 59.00          & 24.65          & 28.80          & 60.17          & 55.23          & 48.46          & 43.67          \\
                      & FLAP          & 36.52          & 61.99          & 70.80 & 78.00          & 48.92          & 34.00          & 70.57          & \underline{74.37} & 64.96          & 60.01          \\
                      & MC-SMoE       & 39.76          & 63.09          & \underline{85.84}          & \underline{83.00} & \underline{56.30} & 39.40          & 74.54          & \textbf{78.34} & \textbf{69.14} & 65.49          \\
                      & RS            & \textbf{48.04} & \textbf{73.06} & \textbf{86.61} & \underline{83.00} & \textbf{61.37} & \textbf{42.40} & \textbf{79.43} & 70.40          & \underline{68.27} & \textbf{68.06} \\
                      & MoNE (Ours)   & \underline{44.62} & \underline{67.05} & 85.41          & \textbf{85.00} & 54.58          & \underline{41.60} & \underline{79.00} & 71.84          & 67.80          & \underline{66.32} \\ \bottomrule
\end{tabular}
}
}
\end{table}

\begin{table}
\footnotesize
\centering
\caption{
    Zero shot performance with 500 calibration samples from Zyda2 dataset. Best results are in \textbf{bold}, and the second best are \underline{underlined}. }
\label{zyda2-500}
\subfloat[OLMoE]{
\resizebox{\textwidth}{!}{
    \begin{tabular}{@{}ccc*{10}{w{c}{0.95cm}}@{}}
\toprule
Pruning ratio         & Model/Method & Arc-c          & Arc-e          & BoolQ          & COPA           & MMLU           & OBQA           & PIQA           & RTE            & Winogrande     & Average        \\ \midrule
0\%                   & OLMoE        & 49.23          & 76.89          & 70.09          & 85.00          & 53.54          & 44.40          & 79.76          & 71.84          & 68.90          & 66.63          \\ \midrule
\multirow{5}{*}{25\%} & Angular      & 32.76          & 61.91          & 61.71          & 74.00          & 23.13          & 37.60          & 71.65          & 53.07          & 55.09          & 52.33          \\
                      & FLAP         & \underline{37.03} & \underline{63.43} & 64.28          & \underline{81.00} & \underline{41.12} & \underline{38.80} & \underline{72.63} & 54.51          & 63.54          & \underline{57.37} \\
                      & MC-SMoE      & 33.36          & 54.46          & \underline{70.03} & \underline{81.00} & 37.05          & 33.80          & 68.17          & \underline{64.62} & \underline{65.19} & 56.41          \\
                      & RS           & 23.81          & 40.91          & 57.92          & 69.00          & 27.79          & 30.20          & 63.71          & 50.18          & 57.38          & 46.77          \\
                      & MoNE (Ours)  & \textbf{41.04} & \textbf{65.66} & \textbf{70.24} & \textbf{87.00} & \textbf{41.21} & \textbf{40.00} & \textbf{76.61} & \textbf{64.98} & \textbf{66.61} & \textbf{61.48} \\ \midrule
\multirow{5}{*}{50\%} & Angular      & \underline{27.22} & \underline{37.50} & 53.91          & 62.00          & 23.96          & 26.60          & 58.27          & 52.35          & 51.85          & 43.74          \\
\bottomrule
\end{tabular}
}
}

\subfloat[Moonlight]{
\resizebox{\textwidth}{!}{
    \begin{tabular}{@{}ccc*{10}{w{c}{0.95cm}}@{}}
\toprule
Pruning ratio         & Model/Method & Arc-c          & Arc-e          & BoolQ          & COPA           & MMLU           & OBQA           & PIQA           & RTE            & Winogrande     & Average        \\ \midrule
0\%                   & Moonlight    & 58.28          & 82.49          & 80.40          & 92.00          & 67.30          & 45.60          & 81.12          & 65.70          & 71.11          & 71.56          \\ \midrule
\multirow{5}{*}{25\%} & Angular      & 39.76          & 52.69          & 38.90          & 79.00          & 42.57          & 32.20          & 68.50          & \underline{61.01} & 62.04          & 52.96          \\
                      & FLAP         & 48.55          & 76.05          & 77.49          & 89.00          & \textbf{55.12} & 42.40          & 76.66          & \textbf{65.34} & 68.59          & 66.58          \\
                      & MC-SMoE      & 37.46          & 63.47          & 76.82          & 81.00          & 48.27          & 35.00          & 71.11          & 58.84          & 70.17          & 60.24          \\
                      & RS           & \textbf{55.63} & \textbf{79.46} & \underline{78.93} & \textbf{91.00} & 46.60          & \textbf{45.80} & \textbf{80.90} & 59.93          & \textbf{72.14} & \underline{67.82} \\
                      & MoNE (Ours)  & \underline{55.03} & \underline{78.96} & \textbf{79.36} & \underline{90.00} & \underline{54.39} & \underline{45.40} & \underline{80.69} & 58.48          & \underline{71.74} & \textbf{68.23} \\ \midrule
\multirow{5}{*}{50\%} & Angular      & 27.90          & 28.54          & 48.01          & 49.00          & \underline{25.67} & 28.80          & 52.56          & 51.99          & 47.75          & 40.02          \\
\bottomrule
\end{tabular}
}
} 

\subfloat[Deepseek-V2-Lite]{
\resizebox{\textwidth}{!}{
    \begin{tabular}{@{}ccc*{10}{w{c}{0.95cm}}@{}}
\toprule
Pruning ratio         & Model/Method     & Arc-c          & Arc-e          & BoolQ          & COPA           & MMLU           & OBQA           & PIQA           & RTE            & Winogrande     & Average        \\ \midrule
0\%                   & Deepseek-V2-Lite & 48.72          & 76.18          & 79.88          & 88.00          & 54.96          & 43.60          & 80.25          & 61.37          & 71.51          & 67.16          \\ \midrule
\multirow{5}{*}{25\%} & Angular          & 32.00          & 53.28          & 64.92          & 75.00          & 26.95          & 34.00          & 71.33          & 58.84          & 61.01          & 53.04          \\
                      & FLAP             & 43.86          & 72.18          & \underline{75.93} & 85.00          & \underline{47.22} & \underline{41.80} & 78.45          & \textbf{62.09} & 68.27          & 63.87          \\
                      & MC-SMoE          & 33.53          & 52.95          & 73.67          & 81.00          & 41.68          & 32.20          & 66.70          & 52.35          & 70.17          & 56.03          \\
                      & RS               & \textbf{48.98} & \textbf{73.23} & 71.77          & \underline{89.00} & 52.68          & \textbf{44.60} & \underline{79.33} & \underline{61.01} & \underline{70.32} & \textbf{65.66} \\
                      & MoNE (Ours)      & \underline{44.62} & \underline{73.11} & \textbf{78.01} & \textbf{90.00} & \textbf{48.29} & \underline{41.80} & \textbf{79.43} & 59.21          & \textbf{71.35} & \underline{65.09} \\ \midrule
\multirow{5}{*}{50\%} & Angular          & 24.06          & 32.79          & 40.40          & 61.00          & 23.22          & 26.80          & 56.42          & \underline{57.76} & 49.09          & 41.28          \\
\bottomrule
\end{tabular}
}
}
\end{table}

\begin{table}
\footnotesize
\centering
\caption{
    Zero shot performance with 1000 calibration samples from Zyda2 dataset. Best results are in \textbf{bold}, and the second best are \underline{underlined}. }
\label{zyda2-1000}
\subfloat[OLMoE]{
\resizebox{\textwidth}{!}{
    \begin{tabular}{@{}ccc*{10}{w{c}{0.95cm}}@{}}
\toprule
Pruning ratio         & Model/Method & Arc-c          & Arc-e          & BoolQ          & COPA           & MMLU           & OBQA           & PIQA           & RTE            & Winogrande     & Average        \\ \midrule
0\%                   & OLMoE        & 49.23          & 76.89          & 70.09          & 85.00          & 53.54          & 44.40          & 79.76          & 71.84          & 68.90          & 66.63          \\ \midrule
\multirow{5}{*}{25\%} & Angular      & 32.76          & 61.91          & 61.71          & 74.00          & 23.13          & \underline{37.60} & 71.65          & 53.07          & 55.09          & 52.33          \\
                      & FLAP         & \underline{38.91} & \textbf{66.20} & \underline{64.65} & \underline{79.00} & \underline{40.05} & \underline{37.60} & \underline{74.65} & \underline{62.82} & 63.77          & \underline{58.63} \\
                      & MC-SMoE      & 38.31          & 61.66          & 61.87          & 73.00          & 33.85          & 33.20          & 66.38          & 57.04          & \underline{65.43} & 54.53          \\
                      & RS           & 26.62          & 43.35          & 59.76          & 70.00          & 27.50          & 30.00          & 65.02          & 49.46          & 56.67          & 47.60          \\
                      & MoNE (Ours)  & \textbf{42.32} & \underline{64.52} & \textbf{66.45} & \textbf{88.00} & \textbf{41.55} & \textbf{40.80} & \textbf{78.02} & \textbf{64.26} & \textbf{67.01} & \textbf{61.44} \\ \midrule
\multirow{5}{*}{50\%} & Angular      & \underline{27.22} & \underline{37.50} & 53.91          & 62.00          & 23.96          & 26.60          & 58.27          & 52.35          & 51.85          & 43.74          \\
\bottomrule
\end{tabular}
}
}

\subfloat[Moonlight]{
\resizebox{\textwidth}{!}{
    \begin{tabular}{@{}ccc*{10}{w{c}{0.95cm}}@{}}
\toprule
Pruning ratio         & Model/Method & Arc-c          & Arc-e          & BoolQ          & COPA           & MMLU           & OBQA           & PIQA           & RTE            & Winogrande     & Average        \\ \midrule
0\%                   & Moonlight    & 58.28          & 82.49          & 80.40          & 92.00          & 67.30          & 45.60          & 81.12          & 65.70          & 71.11          & 71.56          \\ \midrule
\multirow{5}{*}{25\%} & Angular      & 39.76          & 52.69          & 38.90          & 79.00          & 42.57          & 32.20          & 68.50          & \underline{61.01} & 62.04          & 52.96          \\
                      & FLAP         & 49.49          & 76.52          & 76.91          & \underline{90.00} & \underline{54.47} & 42.00          & 77.26          & \textbf{65.70} & \underline{69.06}          & 66.82          \\
                      & MC-SMoE      & 34.81          & 57.74          & 77.22          & 85.00          & 36.13          & 34.60          & 71.49          & 58.12          & 71.35          & 58.50          \\
                      & RS           & \textbf{56.23} & \underline{79.25} & \underline{79.02} & \textbf{91.00} & 46.76          & \underline{45.20} & \underline{80.63} & 59.57          & \textbf{72.14} & \underline{67.76} \\
                      & MoNE (Ours)  & \underline{55.55} & \textbf{79.29} & \textbf{79.66} & \underline{90.00} & \textbf{54.79} & \textbf{45.60} & \textbf{80.69} & 59.21          & \textbf{72.14} & \textbf{68.55} \\ \midrule
\multirow{5}{*}{50\%} & Angular      & 27.90          & 28.54          & 48.01          & 49.00          & \underline{25.67} & 28.80          & 52.56          & 51.99          & 47.75          & 40.02          \\
\bottomrule
\end{tabular}
}
} 

\subfloat[Deepseek-V2-Lite]{
\resizebox{\textwidth}{!}{
    \begin{tabular}{@{}ccc*{10}{w{c}{0.95cm}}@{}}
\toprule
Pruning ratio         & Model/Method     & Arc-c          & Arc-e          & BoolQ          & COPA           & MMLU           & OBQA           & PIQA           & RTE            & Winogrande     & Average        \\ \midrule
0\%                   & Deepseek-V2-Lite & 48.72          & 76.18          & 79.88          & 88.00          & 54.96          & 43.60          & 80.25          & 61.37          & 71.51          & 67.16          \\ \midrule
\multirow{5}{*}{25\%} & Angular          & 32.00          & 53.28          & 64.92          & 75.00          & 26.95          & 34.00          & 71.33          & 58.84          & 61.01          & 53.04          \\
                      & FLAP             & 44.28          & 72.39          & \underline{76.12} & 85.00          & 47.16          & 41.20          & 78.29          & \textbf{62.82} & 67.40          & 63.85          \\
                      & MC-SMoE          & 39.33          & 64.31          & 71.53          & 85.00          & 41.84          & 40.00          & 76.12          & 58.48          & 69.53          & 60.68          \\
                      & RS               & \textbf{49.23} & \underline{73.32} & 71.38          & \textbf{89.00} & \textbf{52.44} & \textbf{45.00} & \textbf{79.71} & \underline{60.29} & \textbf{71.19} & \textbf{65.73} \\
                      & MoNE (Ours)      & \underline{44.71} & \textbf{73.99} & \textbf{78.35} & \underline{88.00} & \underline{49.19} & \underline{42.20} & \underline{79.27} & 59.93          & \underline{70.88} & \underline{65.17} \\ \midrule
\multirow{5}{*}{50\%} & Angular          & 24.06          & 32.79          & 40.40          & 61.00          & 23.22          & 26.80          & 56.42          & \textbf{57.76} & 49.09          & 41.28          \\
\bottomrule
\end{tabular}
}
}
\end{table}

\begin{table}
\footnotesize
\centering
\caption{
    Zero shot performance with 100 calibration samples from C4 dataset. Best results are in \textbf{bold}, and the second best are \underline{underlined}. }
\label{c4-100}
\subfloat[OLMoE]{
\resizebox{\textwidth}{!}{
    \begin{tabular}{@{}ccc*{10}{w{c}{0.95cm}}@{}}
\toprule
Pruning ratio         & Model/Method & Arc-c          & Arc-e          & BoolQ          & COPA           & MMLU           & OBQA           & PIQA           & RTE            & Winogrande     & Average        \\ \midrule
0\%                   & OLMoE        & 49.23          & 76.89          & 70.09          & 85.00          & 53.54          & 44.40          & 79.76          & 71.84          & 68.90          & 66.63          \\ \midrule
\multirow{5}{*}{25\%} & Angular      & 32.76          & \underline{61.91} & 61.71          & 74.00          & 23.13          & 37.60          & 71.65          & 53.07          & 55.09          & 52.33          \\
                      & FLAP         & \underline{36.01} & \textbf{62.67} & \underline{64.83} & 75.00          & \textbf{36.31} & 37.00          & 75.73          & \textbf{58.84} & 62.75          & \underline{56.57} \\
                      & MC-SMoE      & 32.76          & 51.05          & 54.71          & 80.00          & 23.05          & 37.80          & 70.89          & 53.43          & \underline{66.61} & 52.26          \\
                      & RS           & 34.56          & 50.38          & 63.64          & \underline{85.00} & \underline{26.51} & \underline{39.60} & \underline{76.01} & 55.96          & 64.01          & 55.07          \\
                      & MoNE (Ours)  & \textbf{36.69} & 55.18          & \textbf{67.03} & \textbf{86.00} & 25.53          & \textbf{40.20} & \textbf{77.80} & \underline{56.68} & \textbf{67.64} & \textbf{56.97} \\ \midrule
\multirow{5}{*}{50\%} & Angular      & \textbf{27.22} & 37.50          & 53.91          & \underline{62.00} & \underline{23.96} & 26.60          & 58.27          & \underline{52.35} & 51.85          & 43.74          \\
\bottomrule
\end{tabular}
}
}

\subfloat[Moonlight]{
\resizebox{\textwidth}{!}{
    \begin{tabular}{@{}ccc*{10}{w{c}{0.95cm}}@{}}
\toprule
Pruning ratio         & Model/Method & Arc-c          & Arc-e          & BoolQ          & COPA           & MMLU           & OBQA           & PIQA           & RTE            & Winogrande     & Average        \\ \midrule
0\%                   & Moonlight    & 58.28          & 82.49          & 80.40          & 92.00          & 67.30          & 45.60          & 81.12          & 65.70          & 71.11          & 71.56          \\ \midrule
\multirow{5}{*}{25\%} & Angular      & 39.76          & 52.69          & 38.90          & 79.00          & 42.57          & 32.20          & 68.50          & \underline{61.01} & 62.04          & 52.96          \\
                      & FLAP         & 46.76          & 75.04          & 75.90          & 85.00          & \underline{52.98} & 41.00          & 78.07          & \textbf{65.70} & 67.40          & 65.32          \\
                      & MC-SMoE      & 48.38          & 74.20          & \underline{78.69} & \textbf{92.00} & 50.41          & 44.20          & 81.18          & 54.87          & 71.19          & 66.12          \\
                      & RS           & \textbf{55.97} & \textbf{79.76} & \textbf{78.93} & \underline{91.00} & 52.74          & \textbf{46.20} & \textbf{81.39} & 59.21          & \textbf{72.38} & \textbf{68.62} \\
                      & MoNE (Ours)  & \underline{54.27} & \underline{79.25} & 78.07          & 90.00          & \textbf{53.78} & \underline{45.60} & \underline{81.34} & 57.76          & \underline{72.14} & \underline{68.02} \\ \midrule
\multirow{5}{*}{50\%} & Angular      & 27.90          & 28.54          & 48.01          & 49.00          & \underline{25.67} & 28.80          & 52.56          & 51.99          & 47.75          & 40.02          \\
\bottomrule
\end{tabular}
}
} 

\subfloat[Deepseek-V2-Lite]{
\resizebox{\textwidth}{!}{
    \begin{tabular}{@{}ccc*{10}{w{c}{0.95cm}}@{}}
\toprule
Pruning ratio         & Model/Method     & Arc-c          & Arc-e          & BoolQ          & COPA           & MMLU           & OBQA           & PIQA           & RTE            & Winogrande     & Average        \\ \midrule
0\%                   & Deepseek-V2-Lite & 48.72          & 76.18          & 79.88          & 88.00          & 54.96          & 43.60          & 80.25          & 61.37          & 71.51          & 67.16          \\ \midrule
\multirow{5}{*}{25\%} & Angular          & 32.00          & 53.28          & 64.92          & 75.00          & 26.95          & 34.00          & 71.33          & 58.84          & 61.01          & 53.04          \\
                      & FLAP             & 39.68          & 66.58          & \underline{76.45} & \underline{84.00} & 36.14          & 39.80          & 78.67          & 57.04          & 67.48          & 60.65          \\
                      & MC-SMoE          & 36.52          & 59.30          & 59.76          & 81.00          & 36.80          & 37.20          & 75.30          & 54.87          & \underline{69.22} & \underline{56.66} \\
                      & RS               & \textbf{49.32} & \textbf{74.41} & 69.39          & \textbf{90.00} & \textbf{50.35} & \textbf{43.80} & \underline{80.14} & \underline{62.09} & \textbf{70.24} & 65.53          \\
                      & MoNE (Ours)      & \underline{47.44} & \underline{74.03} & \textbf{77.71} & \textbf{90.00} & \underline{49.10} & \underline{42.60} & \textbf{80.25} & \textbf{63.18} & \textbf{70.24} & \textbf{66.06} \\ \midrule
\multirow{5}{*}{50\%} & Angular          & 24.06          & 32.79          & 40.40          & 61.00          & 23.22          & 26.80          & 56.42          & 57.76          & 49.09          & 41.28          \\
\bottomrule
\end{tabular}
}
}
\end{table}

\begin{table}
\footnotesize
\centering
\caption{
    Zero shot performance with 500 calibration samples from C4 dataset. Best results are in \textbf{bold}, and the second best are \underline{underlined}. }
\label{c4-500}
\subfloat[OLMoE]{
\resizebox{\textwidth}{!}{
    \begin{tabular}{@{}ccc*{10}{w{c}{0.95cm}}@{}}
\toprule
Pruning ratio         & Model/Method & Arc-c          & Arc-e          & BoolQ          & COPA           & MMLU           & OBQA           & PIQA           & RTE            & Winogrande     & Average        \\ \midrule
0\%                   & OLMoE        & 49.23          & 76.89          & 70.09          & 85.00          & 53.54          & 44.40          & 79.76          & 71.84          & 68.90          & 66.63          \\ \midrule
\multirow{5}{*}{25\%} & Angular      & 32.76          & \underline{61.91} & 61.71          & \underline{74.00} & 23.13          & 37.60          & 71.65          & 53.07          & 55.09          & 52.33          \\
                      & FLAP         & \underline{36.26} & \textbf{63.80} & 63.49          & 72.00          & \underline{35.09} & 36.00          & 75.35          & 53.07          & 61.96          & 55.23          \\
                      & MC-SMoE      & 26.71          & 48.61          & \textbf{65.90} & 69.00          & \textbf{35.14} & 31.00          & 61.92          & 54.87          & 61.09          & 50.47          \\
                      & RS           & 34.56          & 49.41          & 63.85          & \textbf{86.00} & 27.23          & \underline{39.80} & \underline{75.57} & \underline{56.32} & \underline{64.40} & \underline{55.24} \\
                      & MoNE (Ours)  & \textbf{37.88} & 55.89          & \underline{65.23} & \textbf{86.00} & 25.03          & \textbf{41.20} & \textbf{77.42} & \textbf{57.40} & \textbf{67.80} & \textbf{57.10} \\ \midrule
\multirow{5}{*}{50\%} & Angular      & \underline{27.22} & 37.50          & 53.91          & 62.00          & \underline{23.96} & 26.60          & 58.27          & 52.35          & 51.85          & 43.74          \\
\bottomrule
\end{tabular}
}
}

\subfloat[Moonlight]{
\resizebox{\textwidth}{!}{
    \begin{tabular}{@{}ccc*{10}{w{c}{0.95cm}}@{}}
\toprule
Pruning ratio         & Model/Method & Arc-c          & Arc-e          & BoolQ          & COPA           & MMLU           & OBQA           & PIQA           & RTE            & Winogrande     & Average        \\ \midrule
0\%                   & Moonlight    & 58.28          & 82.49          & 80.40          & 92.00          & 67.30          & 45.60          & 81.12          & 65.70          & 71.11          & 71.56          \\ \midrule
\multirow{5}{*}{25\%} & Angular      & 39.76          & 52.69          & 38.90          & 79.00          & 42.57          & 32.20          & 68.50          & \underline{61.01} & 62.04          & 52.96          \\
                      & FLAP         & 45.22          & 73.40          & 72.29          & 84.00          & \textbf{53.81} & 40.40          & 78.13          & \textbf{64.98} & 67.32          & 64.40          \\
                      & MC-SMoE      & 48.38          & 74.16          & \underline{78.56} & 89.00          & 48.78          & 43.00          & \underline{81.12} & 58.48          & \underline{71.03} & 65.84          \\
                      & RS           & \textbf{55.72} & \textbf{79.80} & \textbf{79.20} & \textbf{91.00} & 52.93          & \underline{44.60} & \textbf{81.23} & 59.57          & \textbf{71.74} & \textbf{68.42} \\
                      & MoNE (Ours)  & \underline{54.52} & \underline{78.87} & \underline{78.56} & \underline{90.00} & \underline{53.80} & \textbf{45.20} & 81.07          & 57.04          & \textbf{71.74} & \underline{67.87} \\ \midrule
\multirow{5}{*}{50\%} & Angular      & 27.90          & 28.54          & 48.01          & 49.00          & \underline{25.67} & 28.80          & 52.56          & 51.99          & 47.75          & 40.02          \\
\bottomrule
\end{tabular}
}
} 

\subfloat[Deepseek-V2-Lite]{
\resizebox{\textwidth}{!}{
    \begin{tabular}{@{}ccc*{10}{w{c}{0.95cm}}@{}}
\toprule
Pruning ratio         & Model/Method     & Arc-c          & Arc-e          & BoolQ          & COPA           & MMLU           & OBQA           & PIQA           & RTE            & Winogrande     & Average        \\ \midrule
0\%                   & Deepseek-V2-Lite & 48.72          & 76.18          & 79.88          & 88.00          & 54.96          & 43.60          & 80.25          & 61.37          & 71.51          & 67.16          \\ \midrule
\multirow{5}{*}{25\%} & Angular          & 32.00          & 53.28          & 64.92          & 75.00          & 26.95          & 34.00          & 71.33          & 58.84          & 61.01          & 53.04          \\
                      & FLAP             & 40.53          & 64.81          & \textbf{77.43} & 84.00          & 37.70          & 39.60          & 79.43          & 57.04          & 67.32          & 60.87          \\
                      & MC-SMoE          & 39.33          & 61.99          & 60.95          & 84.00          & 41.37          & 38.60          & 77.75          & 58.48          & 62.04          & 58.28          \\
                      & RS               & \textbf{49.66} & \textbf{74.49} & 65.63          & \textbf{91.00} & \textbf{50.04} & \textbf{43.60} & \textbf{79.92} & \underline{61.01} & \underline{69.77} & \underline{65.01} \\
                      & MoNE (Ours)      & \underline{46.59} & \underline{73.19} & \underline{77.37} & \underline{89.00} & \underline{48.80} & \underline{43.40} & \underline{79.87} & \textbf{62.09} & \textbf{70.80} & \textbf{65.68} \\ \midrule
\multirow{5}{*}{50\%} & Angular          & 24.06          & 32.79          & 40.40          & 61.00          & 23.22          & 26.80          & 56.42          & \textbf{57.76} & 49.09          & 41.28          \\
\bottomrule
\end{tabular}
}
}
\end{table}

\begin{table}
\footnotesize
\centering
\caption{
    Zero shot performance with 1000 calibration samples from C4 dataset. Best results are in \textbf{bold}, and the second best are \underline{underlined}. }
\label{C4-1000}
\subfloat[OLMoE]{
\resizebox{\textwidth}{!}{
    \begin{tabular}{@{}ccc*{10}{w{c}{0.95cm}}@{}}
\toprule
Pruning ratio         & Model/Method & Arc-c          & Arc-e          & BoolQ          & COPA           & MMLU           & OBQA           & PIQA           & RTE            & Winogrande     & Average        \\ \midrule
0\%                   & OLMoE        & 49.23          & 76.89          & 70.09          & 85.00          & 53.54          & 44.40          & 79.76          & 71.84          & 68.90          & 66.63          \\ \midrule
\multirow{5}{*}{25\%} & Angular      & 32.76          & \underline{61.91} & 61.71          & 74.00          & 23.13          & 37.60          & 71.65          & 53.07          & 55.09          & 52.33          \\
                      & FLAP         & 36.35          & 61.24          & 64.50          & 72.00          & \textbf{37.08} & 36.00          & 75.41          & \underline{57.40} & 62.59          & 55.84          \\
                      & MC-SMoE      & \textbf{40.10} & \textbf{68.22} & 61.65          & 70.00          & \underline{36.91} & 39.60          & 71.65          & \textbf{59.57} & 57.93          & \underline{56.18} \\
                      & RS           & 34.39          & 49.41          & \underline{64.89} & \underline{85.00} & 26.64          & \underline{40.60} & \underline{75.95} & 57.04          & \underline{63.46} & 55.26          \\
                      & MoNE (Ours)  & \underline{38.74} & 57.45          & \textbf{65.57} & \textbf{86.00} & 24.59          & \textbf{41.80} & \textbf{77.42} & 56.68          & \textbf{68.43} & \textbf{57.41} \\ \midrule
\multirow{5}{*}{50\%} & Angular      & \underline{27.22} & 37.50          & 53.91          & 62.00          & \underline{23.96} & 26.60          & 58.27          & 52.35          & 51.85          & 43.74          \\
\bottomrule
\end{tabular}
}
}

\subfloat[Moonlight]{
\resizebox{\textwidth}{!}{
    \begin{tabular}{@{}ccc*{10}{w{c}{0.95cm}}@{}}
\toprule
Pruning ratio         & Model/Method & Arc-c          & Arc-e          & BoolQ          & COPA           & MMLU           & OBQA           & PIQA           & RTE            & Winogrande     & Average        \\ \midrule
0\%                   & Moonlight    & 58.28          & 82.49          & 80.40          & 92.00          & 67.30          & 45.60          & 81.12          & 65.70          & 71.11          & 71.56          \\ \midrule
\multirow{5}{*}{25\%} & Angular      & 39.76          & 52.69          & 38.90          & 79.00          & 42.57          & 32.20          & 68.50          & \underline{61.01} & 62.04          & 52.96          \\
                      & FLAP         & 45.22          & 74.20          & 72.78          & 86.00          & \underline{53.40} & 41.20          & 78.35          & \textbf{65.34} & 68.90          & 65.04          \\
                      & MC-SMoE      & 51.62          & 76.52          & 77.86          & \textbf{92.00} & 43.08          & 43.60          & 80.25          & 55.60          & 71.11          & 65.74          \\
                      & RS           & \textbf{55.97} & \textbf{79.59} & \textbf{78.75} & \underline{90.00} & 53.01          & \underline{45.20} & \underline{81.34} & 59.57          & \textbf{71.90} & \textbf{68.37} \\
                      & MoNE (Ours)  & \underline{53.75} & \underline{79.38} & \underline{78.53} & \underline{90.00} & \textbf{53.75} & \textbf{45.80} & \textbf{81.39} & 57.76          & \underline{71.59} & \underline{67.99} \\ \midrule
\multirow{5}{*}{50\%} & Angular      & 27.90          & 28.54          & 48.01          & 49.00          & \underline{25.67} & 28.80          & 52.56          & 51.99          & 47.75          & 40.02          \\
\bottomrule
\end{tabular}
}
} 

\subfloat[Deepseek-V2-Lite]{
\resizebox{\textwidth}{!}{
    \begin{tabular}{@{}ccc*{10}{w{c}{0.95cm}}@{}}
\toprule
Pruning ratio         & Model/Method     & Arc-c          & Arc-e          & BoolQ          & COPA           & MMLU           & OBQA           & PIQA           & RTE            & Winogrande     & Average        \\ \midrule
0\%                   & Deepseek-V2-Lite & 48.72          & 76.18          & 79.88          & 88.00          & 54.96          & 43.60          & 80.25          & 61.37          & 71.51          & 67.16          \\ \midrule
\multirow{5}{*}{25\%} & Angular          & 32.00          & 53.28          & 64.92          & 75.00          & 26.95          & 34.00          & 71.33          & 58.84          & 61.01          & 53.04          \\
                      & FLAP             & 39.33          & 66.50          & \underline{76.33} & 84.00          & 38.14          & 40.60          & 78.84          & 59.93          & 67.88          & 61.28          \\
                      & MC-SMoE          & 37.29          & 58.42          & 65.72          & 81.00          & 37.91          & 39.00          & 76.12          & 55.60          & 67.80          & 57.65          \\
                      & RS               & \textbf{49.23} & \textbf{74.33} & 68.44          & \underline{88.00} & \textbf{51.59} & \textbf{44.40} & \textbf{80.25} & \textbf{62.45} & \underline{69.69} & \underline{65.38} \\
                      & MoNE (Ours)      & \underline{45.99} & \underline{72.98} & \textbf{77.43} & \textbf{89.00} & \underline{48.90} & \underline{42.60} & \underline{79.98} & \underline{61.01} & \textbf{71.59} & \textbf{65.50} \\ \midrule
\multirow{5}{*}{50\%} & Angular          & 24.06          & 32.79          & 40.40          & 61.00          & 23.22          & 26.80          & 56.42          & \textbf{57.76} & 49.09          & 41.28          \\
\bottomrule
\end{tabular}
}
}
\end{table}

This section presents the experiment results on Zyda2 dataset with 100, 500 and 1000 calibration samples in Table \ref{zyda2-100}, Table \ref{zyda2-500} and Table \ref{zyda2-1000}.
Table \ref{c4-100}, Table \ref{c4-500} and Table \ref{C4-1000} presents the experiment results on C4 dataset with 100, 500 and 1000 calibration samples.
The observations are similar to those in Section \ref{effect exp} and Section \ref{calib size}.
Nevertheless, all methods incurs more aggressive performance drop at higher (50\%) pruning ratio.
Such phenomenon is inevitable for all model compression approaches and inspires the development of MoNE to push the performance limits of structured pruning.

\section{Redundant expert visualization} \label{redundant experts}
\begin{figure}[!t]
    \vspace{-3mm}
    \centering
    \includegraphics[width=\textwidth]{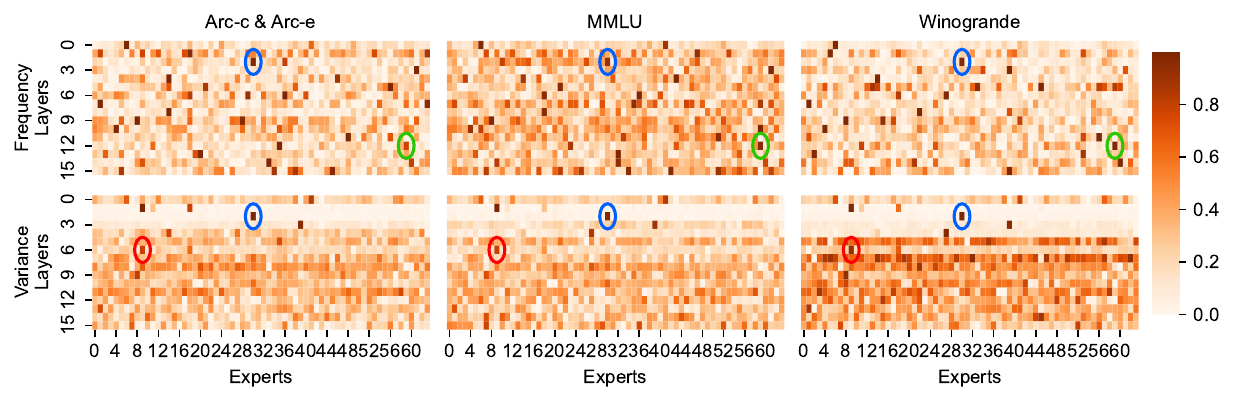}
    \vspace{-6mm}
    \caption{
        Layer-wise normalized expert access frequency and output variance of OLMoE for Arc-C \& Arc-E, MMLU and Winogrande.
    }
    \vspace{-2mm}
    \label{olmoe-1}
\end{figure}

\begin{figure}[!t]
    \vspace{-3mm}
    \centering
    \includegraphics[width=\textwidth]{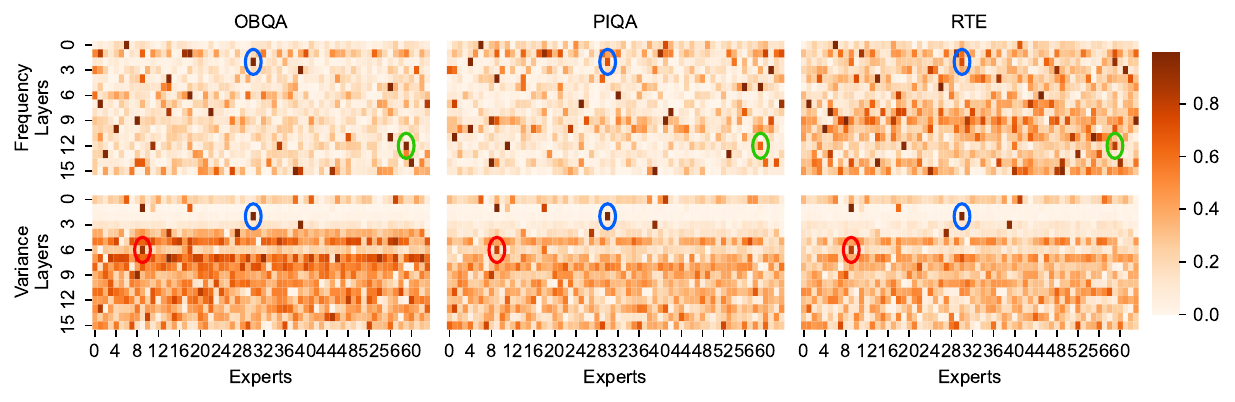}
    \vspace{-6mm}
    \caption{
        Layer-wise normalized expert access frequency and output variance of OLMoE for OBQA, PIQA and RTE.
    }
    \vspace{-2mm}
    \label{olmoe-3}
\end{figure}

\begin{figure}[!t]
    \vspace{-3mm}
    \centering
    \includegraphics[width=\textwidth]{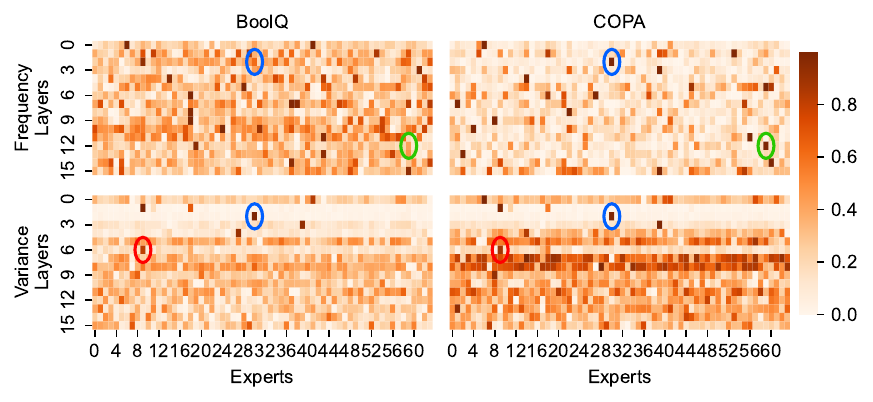}
    \vspace{-6mm}
    \caption{
        Layer-wise normalized expert access frequency and output variance of OLMoE for BoolQ and COPA.
    }
    \vspace{-2mm}
    \label{olmoe-2}
\end{figure}

\begin{figure}[!t]
    \vspace{-3mm}
    \centering
    \includegraphics[width=\textwidth]{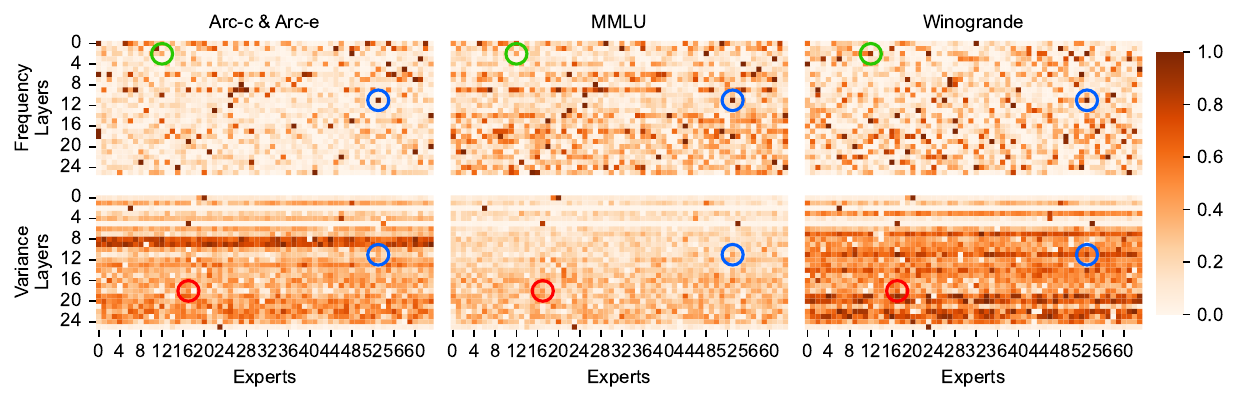}
    \vspace{-6mm}
    \caption{
        Layer-wise normalized expert access frequency and output variance of Moonlight for Arc-C \& Arc-E, MMLU and Winogrande.
    }
    \vspace{-2mm}
    \label{Moonlight-1}
\end{figure}

\begin{figure}[!t]
    \vspace{-3mm}
    \centering
    \includegraphics[width=\textwidth]{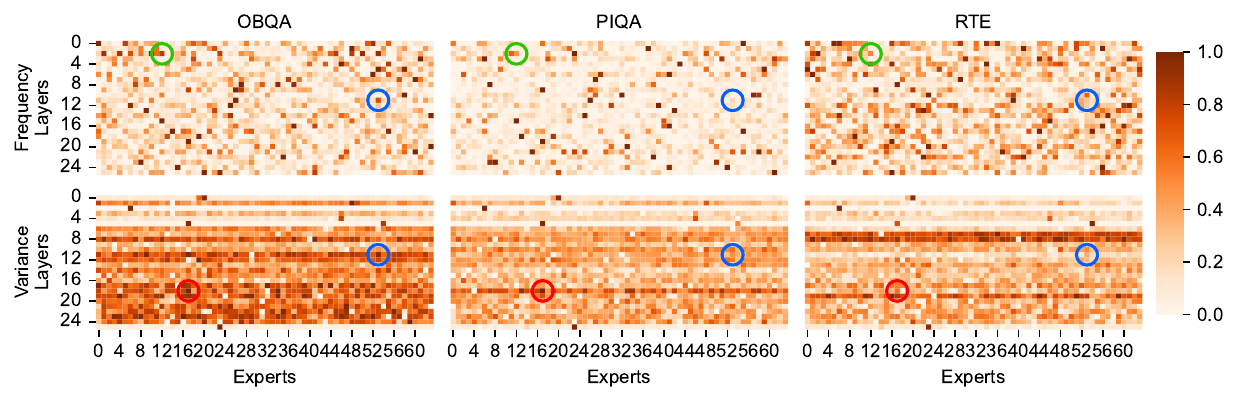}
    \vspace{-6mm}
    \caption{
        Layer-wise normalized expert access frequency and output variance of Moonlight for OBQA, PIQA and RTE.
    }
    \vspace{-2mm}
    \label{Moonlight-3}
\end{figure}

\begin{figure}[!t]
    \vspace{-3mm}
    \centering
    \includegraphics[width=\textwidth]{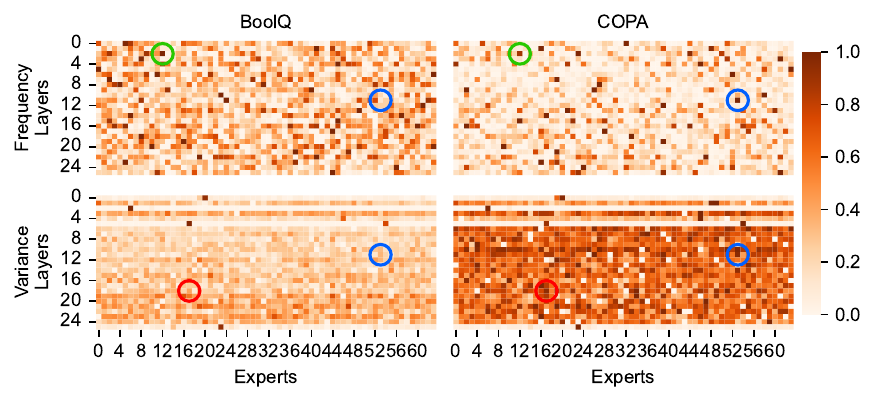}
    \vspace{-6mm}
    \caption{
        Layer-wise normalized expert access frequency and output variance of Moonlight for BoolQ and COPA.
    }
    \vspace{-2mm}
    \label{Moonlight-2}
\end{figure}

\begin{figure}[!t]
    \vspace{-3mm}
    \centering
    \includegraphics[width=\textwidth]{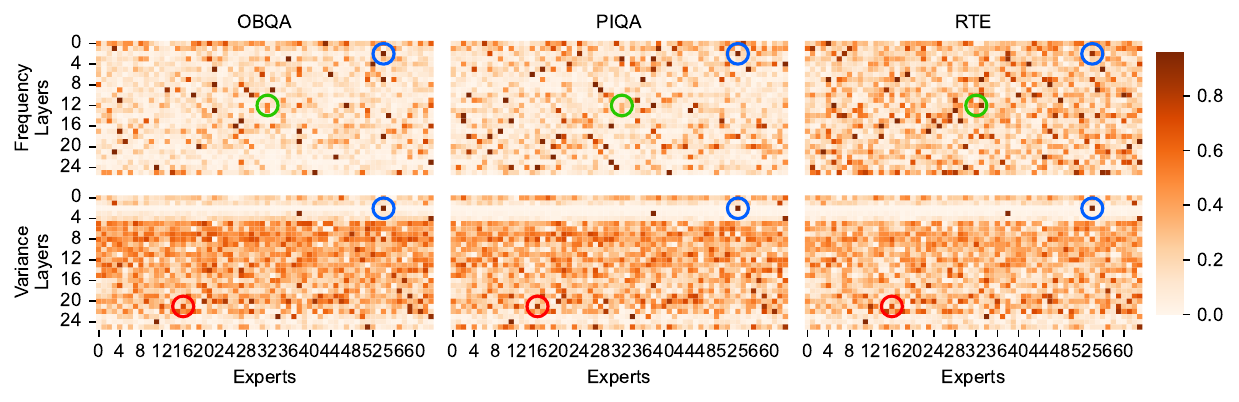}
    \vspace{-6mm}
    \caption{
        Layer-wise normalized expert access frequency and output variance of Deepseek-V2-Lite for OBQA, PIQA and RTE.
    }
    \vspace{-2mm}
    \label{ds-3}
\end{figure}

\begin{figure}[!t]
    \vspace{-3mm}
    \centering
    \includegraphics[width=\textwidth]{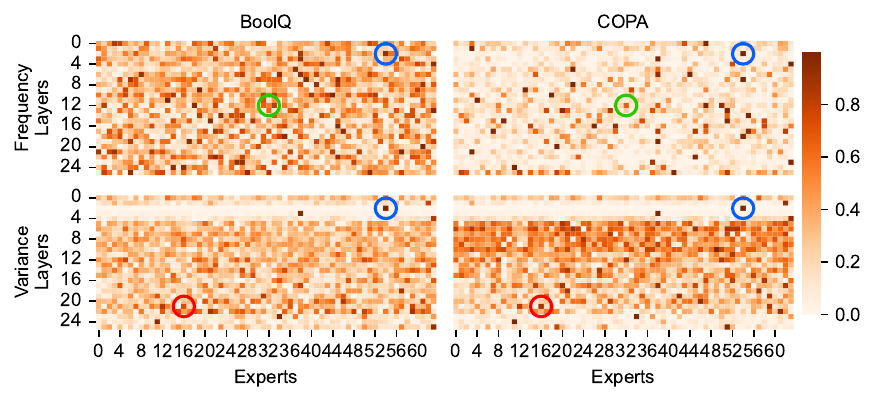}
    \vspace{-6mm}
    \caption{
        Layer-wise normalized expert access frequency and output variance of Deepseek-V2-Lite for BoolQ and COPA.
    }
    \vspace{-2mm}
    \label{ds-2}
\end{figure}
This section complements the visualization of redundant experts for OLMoE, Moonlight and Deepseek-V2-Lite across the nine downstream tasks.
The results are depicted in Figure \ref{olmoe-1} - Figure \ref{ds-2}.
As mentioned in Figure \ref{intro}, for each figure, expert in blue circles has both high frequency and variance. Expert in red circles only has high variance. Expert in green circles only has high frequency.
For each model across the nine downstream tasks, we can always identify the same important experts, validating the effectiveness of the redundancy metric, i.e., the expert access frequency and output variance.

\section{Comprehensive ablation study results} \label{ablation study detail}
\begin{figure}[!t]
    \vspace{-3mm}
    \centering
    \includegraphics[width=\textwidth]{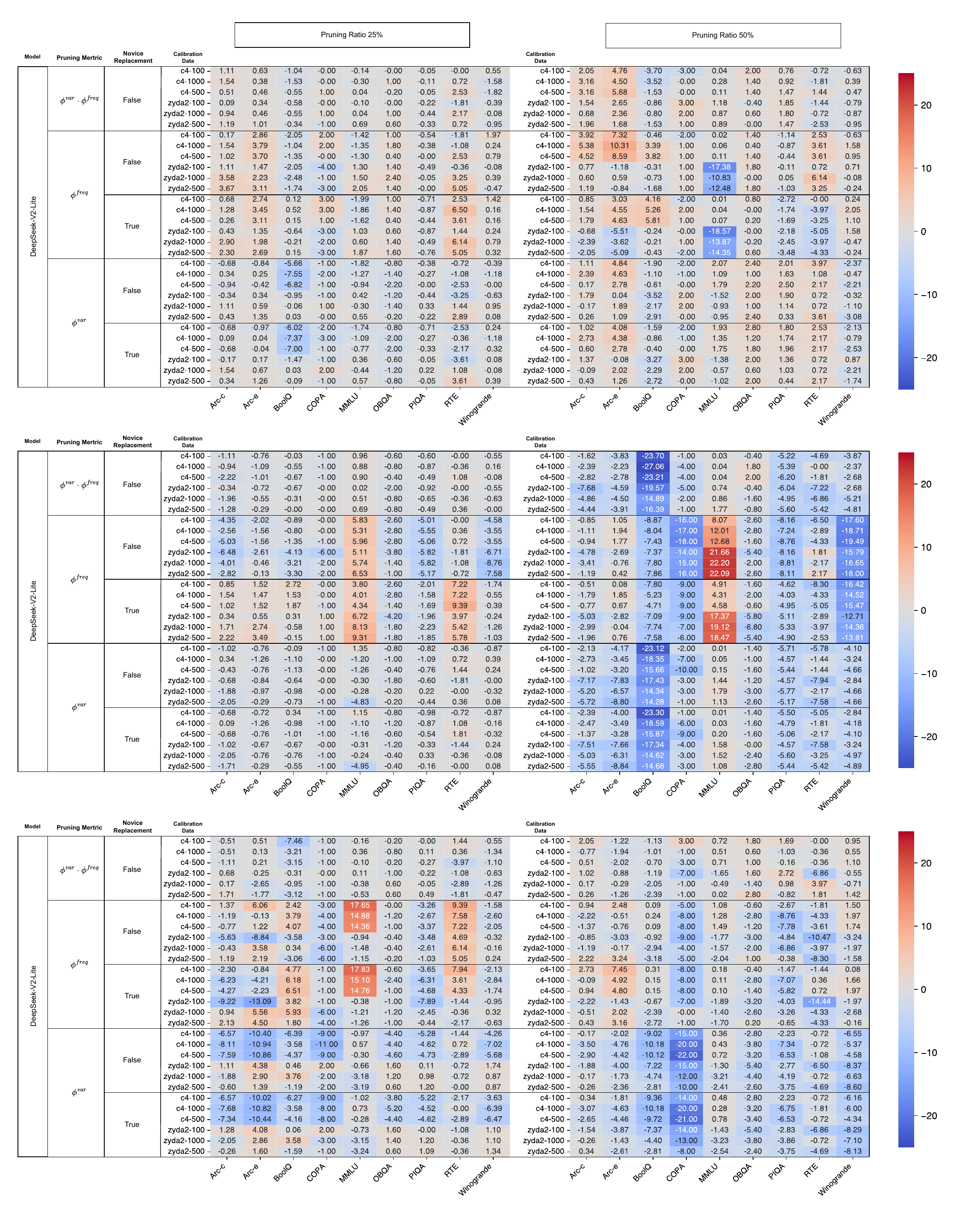}
    \vspace{-6mm}
    \caption{
       Ablation study on expert access frequency, output variance and novice replacement with detailed results.
       Numbers are the difference to the proposed MoNE.
    }
    \vspace{-2mm}
    \label{more ablation}
\end{figure}
Figure \ref{more ablation} reports the detailed ablation study on the impacts of the three factors: expert access frequency, output variance and novice replacement.
The results in this figure validates that the three factors play an important role in maintaining the effectiveness of the pruned models on different model architectures, calibration data sources and calibration sample sizes.
Fusing the three factors ensures the robustness of the proposed MoNE.

\section{Inference latency and memory footprint} \label{deployment}
\begin{table}[t]
\caption{
        Inference latency and memory footprint under different pruning ratio, input sequence and batch size. 
    }
\label{speedup}
\resizebox{\textwidth}{!}{
\begin{tabular}{@{}ccccccc@{}}
\toprule
Pruning ratio & Seed & Batch size & Novice hit ratios* & Total latency (s) & Speedup & Memory usage (GB) \\ \midrule
0\%           & 1020 & 1          & 0                  & 28.86             & 1       & 62.72             \\
25\%          & 1020 & 1          & 0.08               & 27.84             & 1.04    & 47.72             \\
50\%          & 1020 & 1          & 0.35               & 26.27             & 1.10    & 32.73             \\
0\%           & 1998 & 1          & 0                  & 29.01             & 1       & 62.72             \\
25\%          & 1998 & 1          & 0.06               & 28.61             & 1.01    & 47.72             \\
50\%          & 1998 & 1          & 0.34               & 26.22             & 1.11    & 32.73             \\
0\%           & 1020 & 128        & 0                  & 183.12            & 1       & 66.00             \\
25\%          & 1020 & 128        & 0.08               & 174.60            & 1.05    & 51.00             \\
50\%          & 1020 & 128        & 0.34               & 141.07            & 1.30    & 36.01             \\
0\%           & 1998 & 128        & 0                  & 182.37            & 1       & 66.00             \\
25\%          & 1998 & 128        & 0.08               & 169.66            & 1.07    & 51.00             \\
50\%          & 1998 & 128        & 0.33               & 142.38            & 1.28    & 36.01             \\
0\%           & 1020 & 512        & 0                  & 248.02            & 1       & 75.90             \\
25\%          & 1020 & 512        & 0.09               & 212.40            & 1.17    & 60.90             \\
50\%          & 1020 & 512        & 0.33               & 189.57            & 1.31    & 45.91             \\
0\%           & 1998 & 512        & 0                  & 245.42            & 1       & 75.90             \\
25\%          & 1998 & 512        & 0.08               & 212.88            & 1.15    & 60.90             \\
50\%          & 1998 & 512        & 0.34               & 180.92            & 1.36    & 45.91             \\ \bottomrule
\end{tabular}
}
\captionsetup{justification=raggedright, singlelinecheck=false}
\caption*{\footnotesize *Novice hit ratios: the portion of tokens routed to novices across the model.}
\end{table}

While this work mainly focuses on enhancing performance preserving capability of structured pruning, this section evaluates the inference latency and memory footprint for pruned models.
We integrated the pruned Qwen3-30B-A3B with the popular inference framework, SGLang with transformers backend. 
We fixed the 512 input tokens and 256 output tokens, and profiled the latency and memory footprint with SGLang built-in utilities. 
We varied the random seed to generate different input sequences.
Experiments were conducted on a single H20 GPU.
The results are listed in Table \ref{speedup}.

According to Table \ref{speedup}, we would like to clarify three points regarding the latency and memory usage.
First, inference latency speedup of MoNE is sensitive to batch size.
When batch size is 1, there is minor speedup, as the decoding phase only executes one tokens per step, leading to a memory bound situation where the GPU tensor core is underutilized and the major bottleneck is the GPU memory bandwidth \citep{hong2023flashdecoding++, frantar2025marlin}. 
By increasing the batch size to 512, MoNE can achieve speedup of 36\%.
Second, inference latency speedup is sensitive to novice hit ratio instead of novice counts (pruning ratio).
Qwen3-30B-A3B has 128 experts per layer, and each input token is routed to 8 experts per novices. 
The actual speedup is decided by the routing results for each input token at runtime. With higher novice hit ratio, MoNE is expected to attain more speedup. 
While higher pruning ratio may lead to more novices in each layer, it cannot directly translate to speedup. 
For the most extreme case where there is no token routed to novices, the computation overhead (as well as the accuracy) will be the same as the original model.
Finally, Table \ref{speedup} indicates that MoNE achieves maximum memory reduction that consistently increases with the pruning ratio. For a given pruning ratio, the heavy expert parameters are directly replaced by a light weight constant vector.


%% file: ref.bib
@inproceedings{
    gromov2025the,
    title={The Unreasonable Ineffectiveness of the Deeper Layers},
    author={Andrey Gromov and Kushal Tirumala and Hassan Shapourian and Paolo Glorioso and Dan Roberts},
    booktitle={The Thirteenth International Conference on Learning Representations},
    year={2025},
}

@article{ling2024slimgpt,
  title={SlimGPT: Layer-wise Structured Pruning for Large Language Models},
  author={Ling, Gui and Wang, Ziyang and Liu, Qingwen},
  journal={Advances in Neural Information Processing Systems},
  volume={37},
  pages={107112--107137},
  year={2024}
}

@inproceedings{
Fan2020Reducing,
title={Reducing Transformer Depth on Demand with Structured Dropout},
author={Angela Fan and Edouard Grave and Armand Joulin},
booktitle={International Conference on Learning Representations},
year={2020},
}

@article{ma2023llm,
  title={Llm-pruner: On the structural pruning of large language models},
  author={Ma, Xinyin and Fang, Gongfan and Wang, Xinchao},
  journal={Advances in neural information processing systems},
  volume={36},
  pages={21702--21720},
  year={2023}
}

@inproceedings{an2024fluctuation,
  title={Fluctuation-based adaptive structured pruning for large language models},
  author={An, Yongqi and Zhao, Xu and Yu, Tao and Tang, Ming and Wang, Jinqiao},
  booktitle={Proceedings of the AAAI Conference on Artificial Intelligence},
  volume={38},
  number={10},
  pages={10865--10873},
  year={2024}
}

@inproceedings{voita-etal-2019-analyzing,
    title = "Analyzing Multi-Head Self-Attention: Specialized Heads Do the Heavy Lifting, the Rest Can Be Pruned",
    author = "Voita, Elena  and
      Talbot, David  and
      Moiseev, Fedor  and
      Sennrich, Rico  and
      Titov, Ivan",
    editor = "Korhonen, Anna  and
      Traum, David  and
      M{\`a}rquez, Llu{\'i}s",
    booktitle = "Proceedings of the 57th Annual Meeting of the Association for Computational Linguistics",
    month = jul,
    year = "2019",
    address = "Florence, Italy",
    publisher = "Association for Computational Linguistics",
    doi = "10.18653/v1/P19-1580",
    pages = "5797--5808",
}

@article{he2024demystifying,
  title={Demystifying the compression of mixture-of-experts through a unified framework},
  author={He, Shwai and Dong, Daize and Ding, Liang and Li, Ang},
  journal={arXiv preprint arXiv:2406.02500},
  year={2024}
}

@article{muralidharan2024compact,
  title = {Compact language models via pruning and knowledge distillation},
  author = {Muralidharan, Saurav and Turuvekere Sreenivas, Sharath and Joshi, Raviraj and Chochowski, Marcin and Patwary, Mostofa and Shoeybi, Mohammad and Catanzaro, Bryan and Kautz, Jan and Molchanov, Pavlo},
  journal = {Advances in Neural Information Processing Systems},
  volume = {37},
  pages = {41076--41102},
  year = {2024}
}

@inproceedings{
xia2024sheared,
title={Sheared {LL}a{MA}: Accelerating Language Model Pre-training via Structured Pruning},
author={Mengzhou Xia and Tianyu Gao and Zhiyuan Zeng and Danqi Chen},
booktitle={The Twelfth International Conference on Learning Representations},
year={2024},
}

@inproceedings{lu2024not,
  title={Not All Experts are Equal: Efficient Expert Pruning and Skipping for Mixture-of-Experts Large Language Models},
  author={Lu, Xudong and Liu, Qi and Xu, Yuhui and Zhou, Aojun and Huang, Siyuan and Zhang, Bo and Yan, Junchi and Li, Hongsheng},
  booktitle={Proceedings of the 62nd Annual Meeting of the Association for Computational Linguistics (Volume 1: Long Papers)},
  pages={6159--6172},
  year={2024}
}

@inproceedings{
li2024merge,
title={Merge, Then Compress: Demystify Efficient {SM}oE with Hints from Its Routing Policy},
author={Pingzhi Li and Zhenyu Zhang and Prateek Yadav and Yi-Lin Sung and Yu Cheng and Mohit Bansal and Tianlong Chen},
booktitle={The Twelfth International Conference on Learning Representations},
year={2024},
}

@inproceedings{frantar2023sparsegpt,
  title={Sparsegpt: Massive language models can be accurately pruned in one-shot},
  author={Frantar, Elias and Alistarh, Dan},
  booktitle={International Conference on Machine Learning},
  pages={10323--10337},
  year={2023},
  organization={PMLR}
}

@inproceedings{
sun2024a,
title={A Simple and Effective Pruning Approach for Large Language Models},
author={Mingjie Sun and Zhuang Liu and Anna Bair and J Zico Kolter},
booktitle={The Twelfth International Conference on Learning Representations},
year={2024},
}

@inproceedings{
bai2024sparsellm,
title={Sparse{LLM}: Towards Global Pruning of Pre-trained Language Models},
author={Guangji Bai and Yijiang Li and Chen Ling and Kibaek Kim and Liang Zhao},
booktitle={The Thirty-eighth Annual Conference on Neural Information Processing Systems},
year={2024},
}

@inproceedings{
lepikhin2021gshard,
title={{\{}GS{\}}hard: Scaling Giant Models with Conditional Computation and Automatic Sharding},
author={Dmitry Lepikhin and HyoukJoong Lee and Yuanzhong Xu and Dehao Chen and Orhan Firat and Yanping Huang and Maxim Krikun and Noam Shazeer and Zhifeng Chen},
booktitle={International Conference on Learning Representations},
year={2021},
}

@article{liu2024deepseek,
  title={Deepseek-v3 technical report},
  author={Liu, Aixin and Feng, Bei and Xue, Bing and Wang, Bingxuan and Wu, Bochao and Lu, Chengda and Zhao, Chenggang and Deng, Chengqi and Zhang, Chenyu and Ruan, Chong and others},
  journal={arXiv preprint arXiv:2412.19437},
  year={2024}
}

@misc{deepseekv2,
      title={DeepSeek-V2: A Strong, Economical, and Efficient Mixture-of-Experts Language Model}, 
      author={DeepSeek-AI},
      year={2024},
      eprint={2405.04434},
      archivePrefix={arXiv},
      primaryClass={cs.CL}
}

@article{liu2025muon,
  title={Muon is scalable for llm training},
  author={Liu, Jingyuan and Su, Jianlin and Yao, Xingcheng and Jiang, Zhejun and Lai, Guokun and Du, Yulun and Qin, Yidao and Xu, Weixin and Lu, Enzhe and Yan, Junjie and others},
  journal={arXiv preprint arXiv:2502.16982},
  year={2025}
}

@inproceedings{
muennighoff2024olmoe,
title={{OLM}oE: Open Mixture-of-Experts Language Models},
author={Niklas Muennighoff and Luca Soldaini and Dirk Groeneveld and Kyle Lo and Jacob Morrison and Sewon Min and Weijia Shi and Evan Pete Walsh and Oyvind Tafjord and Nathan Lambert and Yuling Gu and Shane Arora and Akshita Bhagia and Dustin Schwenk and David Wadden and Alexander Wettig and Binyuan Hui and Tim Dettmers and Douwe Kiela and Ali Farhadi and Noah A. Smith and Pang Wei Koh and Amanpreet Singh and Hannaneh Hajishirzi},
booktitle={The Thirteenth International Conference on Learning Representations},
year={2025},
}

@inproceedings{
huang2025mixture,
title={Mixture Compressor for Mixture-of-Experts {LLM}s Gains More},
author={Wei Huang and Yue Liao and Jianhui Liu and Ruifei He and Haoru Tan and Shiming Zhang and Hongsheng Li and Si Liu and XIAOJUAN QI},
booktitle={The Thirteenth International Conference on Learning Representations},
year={2025},
}

@misc{zyphra_nvidia_2024,
    author = {Yury Tokpanov and Paolo Glorioso and Ayush Dattagupta and Vibhu Jawa and Ryan Wolf and Vikranth Jeyakumar and Arham Mehta and Quentin Anthony and Beren Millidge},
    title = {Building {Zyda-2}, a 5 {Trillion} {Token} {High-Quality} {Dataset}, with {NVIDIA} {NeMo} {Curator}},
    publisher = {Zyphra},
    year = {2024},
    month = {October},
    day = {15}
}

@article{raffel2020exploring,
  title={Exploring the limits of transfer learning with a unified text-to-text transformer},
  author={Raffel, Colin and Shazeer, Noam and Roberts, Adam and Lee, Katherine and Narang, Sharan and Matena, Michael and Zhou, Yanqi and Li, Wei and Liu, Peter J},
  journal={Journal of machine learning research},
  volume={21},
  number={140},
  pages={1--67},
  year={2020}
}

@misc{eval-harness,
  author       = {Gao, Leo and Tow, Jonathan and Abbasi, Baber and Biderman, Stella and Black, Sid and DiPofi, Anthony and Foster, Charles and Golding, Laurence and Hsu, Jeffrey and Le Noac'h, Alain and Li, Haonan and McDonell, Kyle and Muennighoff, Niklas and Ociepa, Chris and Phang, Jason and Reynolds, Laria and Schoelkopf, Hailey and Skowron, Aviya and Sutawika, Lintang and Tang, Eric and Thite, Anish and Wang, Ben and Wang, Kevin and Zou, Andy},
  title        = {A framework for few-shot language model evaluation},
  month        = 07,
  year         = 2024,
  publisher    = {Zenodo},
  version      = {v0.4.3},
}

@article{clark2018think,
  title={Think you have solved question answering? try arc, the ai2 reasoning challenge},
  author={Clark, Peter and Cowhey, Isaac and Etzioni, Oren and Khot, Tushar and Sabharwal, Ashish and Schoenick, Carissa and Tafjord, Oyvind},
  journal={arXiv preprint arXiv:1803.05457},
  year={2018}
}

@inproceedings{clark2019boolq,
  title={BoolQ: Exploring the Surprising Difficulty of Natural Yes/No Questions},
  author={Clark, Christopher and Lee, Kenton and Chang, Ming-Wei and Kwiatkowski, Tom and Collins, Michael and Toutanova, Kristina},
  booktitle={Proceedings of the 2019 Conference of the North American Chapter of the Association for Computational Linguistics: Human Language Technologies, Volume 1 (Long and Short Papers)},
  pages={2924--2936},
  year={2019}
}

@inproceedings{roemmele2011choice,
  title={Choice of Plausible Alternatives: An Evaluation of Commonsense Causal Reasoning.},
  author={Roemmele, Melissa and Bejan, Cosmin Adrian and Gordon, Andrew S},
  booktitle={AAAI spring symposium: logical formalizations of commonsense reasoning},
  pages={90--95},
  year={2011}
}

@inproceedings{
hendrycks2021measuring,
title={Measuring Massive Multitask Language Understanding},
author={Dan Hendrycks and Collin Burns and Steven Basart and Andy Zou and Mantas Mazeika and Dawn Song and Jacob Steinhardt},
booktitle={International Conference on Learning Representations},
year={2021},
}

@inproceedings{mihaylov2018can,
  title={Can a Suit of Armor Conduct Electricity? A New Dataset for Open Book Question Answering},
  author={Mihaylov, Todor and Clark, Peter and Khot, Tushar and Sabharwal, Ashish},
  booktitle={Proceedings of the 2018 Conference on Empirical Methods in Natural Language Processing},
  pages={2381--2391},
  year={2018}
}

@inproceedings{bisk2020piqa,
  title={Piqa: Reasoning about physical commonsense in natural language},
  author={Bisk, Yonatan and Zellers, Rowan and Gao, Jianfeng and Choi, Yejin and others},
  booktitle={Proceedings of the AAAI conference on artificial intelligence},
  volume={34},
  number={05},
  pages={7432--7439},
  year={2020}
}

@article{wang2019superglue,
  title={Superglue: A stickier benchmark for general-purpose language understanding systems},
  author={Wang, Alex and Pruksachatkun, Yada and Nangia, Nikita and Singh, Amanpreet and Michael, Julian and Hill, Felix and Levy, Omer and Bowman, Samuel},
  journal={Advances in neural information processing systems},
  volume={32},
  year={2019}
}

@article{sakaguchi2021winogrande,
  title={Winogrande: An adversarial winograd schema challenge at scale},
  author={Sakaguchi, Keisuke and Bras, Ronan Le and Bhagavatula, Chandra and Choi, Yejin},
  journal={Communications of the ACM},
  volume={64},
  number={9},
  pages={99--106},
  year={2021},
  publisher={ACM New York, NY, USA}
}

@article{jiang2024mixtral,
  title={Mixtral of experts},
  author={Jiang, Albert Q and Sablayrolles, Alexandre and Roux, Antoine and Mensch, Arthur and Savary, Blanche and Bamford, Chris and Chaplot, Devendra Singh and Casas, Diego de las and Hanna, Emma Bou and Bressand, Florian and others},
  journal={arXiv preprint arXiv:2401.04088},
  year={2024}
}

@inproceedings{jouppi2023tpu,
  title={Tpu v4: An optically reconfigurable supercomputer for machine learning with hardware support for embeddings},
  author={Jouppi, Norm and Kurian, George and Li, Sheng and Ma, Peter and Nagarajan, Rahul and Nai, Lifeng and Patil, Nishant and Subramanian, Suvinay and Swing, Andy and Towles, Brian and others},
  booktitle={Proceedings of the 50th annual international symposium on computer architecture},
  pages={1--14},
  year={2023}
}

@article{team2024qwen2,
  title={Qwen2 technical report},
  author={Team Qwen},
  journal={arXiv preprint arXiv:2407.10671},
  year={2024}
}

@article{yang2025qwen3,
  title={Qwen3 technical report},
  author={Yang, An and Li, Anfeng and Yang, Baosong and Zhang, Beichen and Hui, Binyuan and Zheng, Bo and Yu, Bowen and Gao, Chang and Huang, Chengen and Lv, Chenxu and others},
  journal={arXiv preprint arXiv:2505.09388},
  year={2025}
}

@article{sarkar2024revisiting,
  title={Revisiting smoe language models by evaluating inefficiencies with task specific expert pruning},
  author={Sarkar, Soumajyoti and Lausen, Leonard and Cevher, Volkan and Zha, Sheng and Brox, Thomas and Karypis, George},
  journal={arXiv preprint arXiv:2409.01483},
  year={2024}
}

@article{chen2022task,
  title={Task-specific expert pruning for sparse mixture-of-experts},
  author={Chen, Tianyu and Huang, Shaohan and Xie, Yuan and Jiao, Binxing and Jiang, Daxin and Zhou, Haoyi and Li, Jianxin and Wei, Furu},
  journal={arXiv preprint arXiv:2206.00277},
  year={2022}
}

@inproceedings{lightman2023let,
  title={Let's verify step by step},
  author={Lightman, Hunter and Kosaraju, Vineet and Burda, Yuri and Edwards, Harrison and Baker, Bowen and Lee, Teddy and Leike, Jan and Schulman, John and Sutskever, Ilya and Cobbe, Karl},
  booktitle={The Twelfth International Conference on Learning Representations},
  year={2023}
}

@article{lin2025zebralogic,
  title={Zebralogic: On the scaling limits of llms for logical reasoning},
  author={Lin, Bill Yuchen and Bras, Ronan Le and Richardson, Kyle and Sabharwal, Ashish and Poovendran, Radha and Clark, Peter and Choi, Yejin},
  journal={arXiv preprint arXiv:2502.01100},
  year={2025}
}

@article{zhang2025swe,
  title={SWE-bench Goes Live!},
  author={Zhang, Linghao and He, Shilin and Zhang, Chaoyun and Kang, Yu and Li, Bowen and Xie, Chengxing and Wang, Junhao and Wang, Maoquan and Huang, Yufan and Fu, Shengyu and others},
  journal={arXiv preprint arXiv:2505.23419},
  year={2025}
}

@article{hong2023flashdecoding++,
  title={Flashdecoding++: Faster large language model inference on gpus},
  author={Hong, Ke and Dai, Guohao and Xu, Jiaming and Mao, Qiuli and Li, Xiuhong and Liu, Jun and Chen, Kangdi and Dong, Yuhan and Wang, Yu},
  journal={arXiv preprint arXiv:2311.01282},
  year={2023}
}

@inproceedings{frantar2025marlin,
  title={Marlin: Mixed-precision auto-regressive parallel inference on large language models},
  author={Frantar, Elias and Castro, Roberto L and Chen, Jiale and Hoefler, Torsten and Alistarh, Dan},
  booktitle={Proceedings of the 30th ACM SIGPLAN Annual Symposium on Principles and Practice of Parallel Programming},
  pages={239--251},
  year={2025}
}

@article{xie2024moe,
  title={Moe-pruner: Pruning mixture-of-experts large language model using the hints from its router},
  author={Xie, Yanyue and Zhang, Zhi and Zhou, Ding and Xie, Cong and Song, Ziang and Liu, Xin and Wang, Yanzhi and Lin, Xue and Xu, An},
  journal={arXiv preprint arXiv:2410.12013},
  year={2024}
}

@article{li2025slimmoe,
  title={SlimMoE: Structured Compression of Large MoE Models via Expert Slimming and Distillation},
  author={Li, Zichong and Liang, Chen and Zhang, Zixuan and Hong, Ilgee and Kim, Young Jin and Chen, Weizhu and Zhao, Tuo},
  journal={arXiv preprint arXiv:2506.18349},
  year={2025}
}

@article{zhao2026dydit,
  title={DyDiT++: Diffusion Transformers with Timestep and Spatial Dynamics for Efficient Visual Generation},
  author={Zhao, Wangbo and Han, Yizeng and Tang, Jiasheng and Wang, Kai and Luo, Hao and Song, Yibing and Huang, Gao and Wang, Fan and You, Yang},
  journal={IEEE Transactions on Pattern Analysis and Machine Intelligence},
  year={2026},
  publisher={IEEE}
}

@inproceedings{dynamic,
  title={Dynamic diffusion transformer},
  author={Zhao, Wangbo and Han, Yizeng and Tang, Jiasheng and Wang, Kai and Song, Yibing and Huang, Gao and Wang, Fan and You, Yang},
  journal={ICLR},
  year={2025}
}

@article{zhao2025rapid,
  title={RAPID\^{} 3: Tri-Level Reinforced Acceleration Policies for Diffusion Transformer},
  author={Zhao, Wangbo and Han, Yizeng and Tang, Zhiwei and Tang, Jiasheng and Zhou, Pengfei and Wang, Kai and Zhuang, Bohan and Wang, Zhangyang and Wang, Fan and You, Yang},
  journal={arXiv preprint arXiv:2509.22323},
  year={2025}
}
